\documentclass{article}

\usepackage{microtype}
\usepackage{graphicx}
\usepackage{subfigure}
\usepackage{booktabs} 
\usepackage{sidecap}

\usepackage{array}
\usepackage{multirow}
\usepackage{graphicx}
\usepackage{amsmath}
\usepackage{xcolor}
\date{\vspace{-2ex}}
\usepackage{cancel}
\usepackage{amssymb}
\usepackage{amsthm}
\usepackage{times}  
\usepackage{helvet} 
\usepackage{courier}  
\usepackage{caption} 

\usepackage{amsfonts}
\newcommand\ignore[1]{}
\newcommand\toappendix[1]{}
\usepackage{authblk}

\newcommand{\secref}[1]{Section \ref{#1}}
\newcommand{\figref}[1]{Figure \ref{#1}}
\renewcommand{\eqref}[1]{Eq.~(\ref{#1})}
\newcommand\RLGN[0]{RLGN}

\usepackage{hyperref}

\usepackage[accepted]{icml2021}

\icmltitlerunning{Controlling Graph Dynamics with  Reinforcement Learning and Graph Neural Networks}

\begin{document}

\twocolumn[
\icmltitle{Controlling Graph Dynamics with \\ Reinforcement Learning and Graph Neural Networks}



\icmlsetsymbol{equal}{*}

\begin{icmlauthorlist}
\icmlauthor{Eli A. Meirom}{NVIDIA}
\icmlauthor{Haggai Maron}{NVIDIA}
\icmlauthor{Shie Mannor}{NVIDIA}
\icmlauthor{Gal Chechik}{NVIDIA}
\end{icmlauthorlist}
\icmlaffiliation{NVIDIA}{NVIDIA Research, Israel}
\icmlcorrespondingauthor{Eli Meirom}{emeirom@nvidia.com}

\icmlkeywords{Machine Learning, ICML, Reinforcement Learning, Graph neural networks, epidemic detection, influence maximization}

\vskip 0.3in

]
\printAffiliationsAndNotice{}



\begin{abstract}
We consider the problem of controlling a partially-observed
dynamic process on a graph by a limited number of interventions. This problem naturally arises in contexts such as scheduling virus tests to curb an epidemic; targeted marketing in order to promote a product; and manually inspecting posts to detect fake news spreading on social networks. 

We formulate this setup as a sequential decision problem over a temporal graph process. In face of an exponential state space, combinatorial action space and partial observability, we design a novel tractable scheme to control dynamical processes on temporal graphs. We successfully apply our approach to two popular problems that fall into our framework: prioritizing which nodes should be tested in order to curb the spread of an epidemic, and influence maximization on a graph.

\end{abstract}

\section{Introduction}
\begin{figure}[t]
    \centering
    \includegraphics[width=0.9\columnwidth]{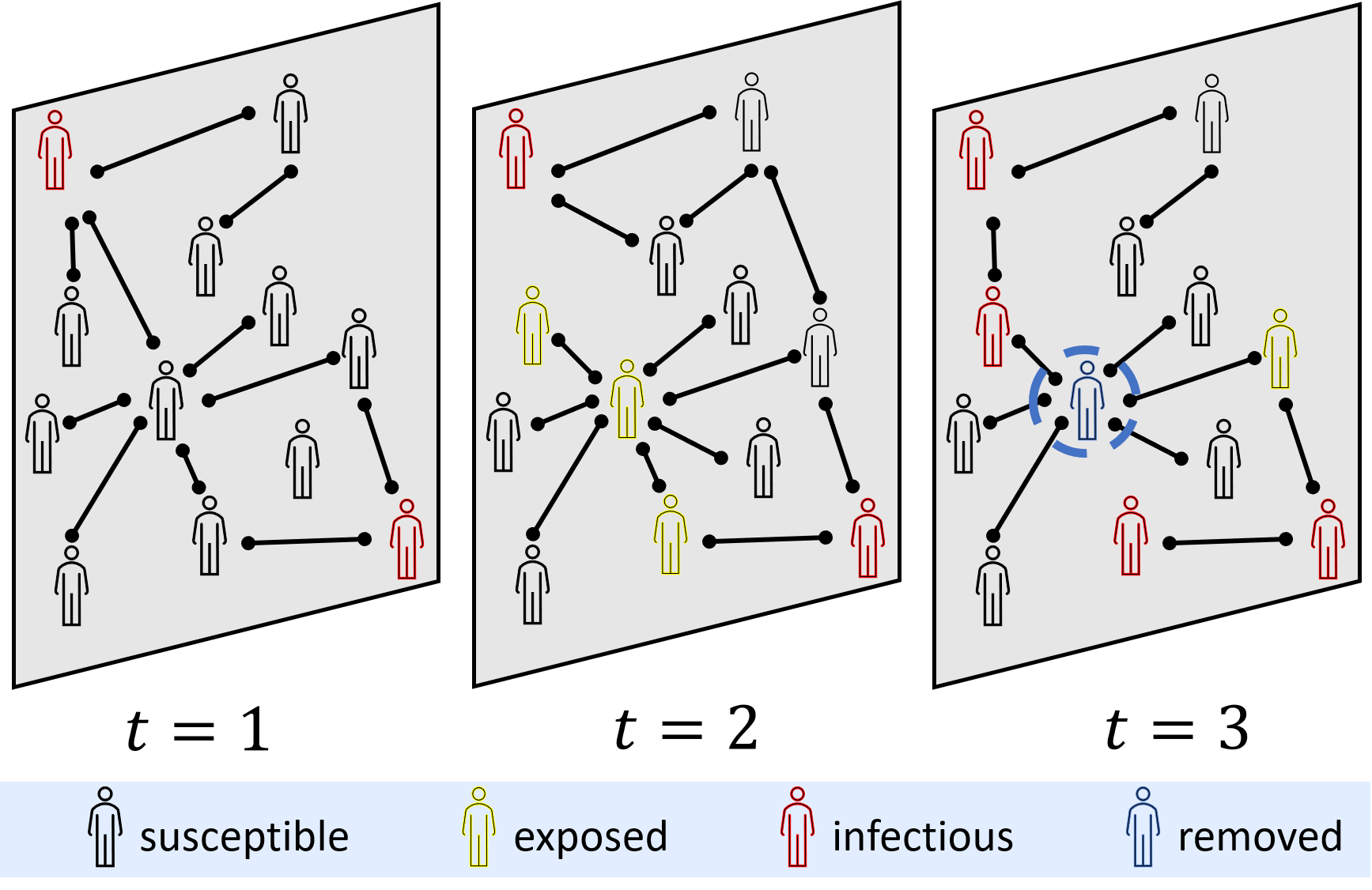}
    \caption{ A viral infection process on a graph and an intervention aimed to stop its spread. Here, graph nodes represent people and edges represent interactions. At $t=1$ only two  people are infected (red). At $t=2$ several interactions resulted in new \textit{exposed} people (yellow); At $t=3$ the blue node was selected to be quarantined to stop the viral spread.  This paper presents a general framework for learning how to control such dynamic processes on graphs. }\label{fig:viralinfection}
\end{figure}

Consider an epidemic spreading in the population. To contain the disease and prevent it from spreading, it becomes critical to detect infected carriers and isolate them; see Fig.~\ref{fig:viralinfection} for an illustration. As the epidemic spreads, the demand for tests outgrows their availability, and not all potential carriers can be tested. It becomes necessary to identify the most likely epidemic carriers using limited testing resources. How should we rank candidates and prioritize vaccines and tests to prevent the disease from spreading? As a second example, imagine a seemingly very different problem, where one would like to promote an opinion or support product adaption by advertisements or information sharing on a social graph.  If an impactful node is convinced, it may influence other nodes towards the desired opinion, creating a cascade of information diffusion. 


These two problems are important examples of a larger family of problems: controlling diffusive processes over networks through nodal interventions. Other examples include viruses inflicting computer networks or cascades of failures in power networks. In all these cases, an agent can steer the dynamics of the system using interventions that modify the states of a (relatively) small number of nodes.  For instance, infected people can be asked to self-quarantine, preventing the spread of a disease, and key twitters may be targeted with coupons. However, a key difficulty is that the current state is often not fully observed, for example, we don't know the ground truth infection status for every node in the graph. 

More formally, we consider a graph $G(t)=(\mathcal{V},\mathcal{E}(t))$
whose structure changes in time. $\mathcal{V}$ is the set of nodes and $\mathcal{E}(t)=\{e_{uv}(t)\}$ is the set of edges at step $t$.  The state of a node $v\in\mathcal{V}$ is a random variable that depends on the interactions between $v$ and its neighbors.  At each turn, the agent may select a subset of nodes and attempt to change their state. The goal is to minimize an objective that depends on the number of nodes in each state. For example, consider a setup where the agent tries to promote its product or opinion. At each step, the agent may select a set of seed nodes and attempt to influence them by presenting relevant information or ads. If those nodes are convinced, they may spread the information through future contacts. The optimization goal, in this case, is to maximize the number of influenced nodes.


The problem of controlling the dynamics of a system using localized interventions is very hard, and for several reasons. First, it requires making decisions in a continuously changing environment with complex dependencies. Second, to solve the problem one must assess the potential downstream ripple effect for any specific node that becomes affected, and balance it with the probability that the node indeed becomes affected. Finally, models must handle noise and partial observability. In particular, it is well known that even the single-round, non-sequential, influence maximization problem is computationally hard \cite{Kempe2003}.

Current approaches for solving this problem can be divided into two main families: (1) Monte Carlo
simulation that estimates the utility of each decision  \citep[see e.g.][]{Goyal2011}. These approaches can find good solutions for small to moderate-sized ($\sim 10^3$ nodes) graphs, but do not scale to larger graphs. (2) Heuristics based on topological properties of the known graph. For example, act on nodes with a high degree \citep[e.g.][]{Liu2017}. These approaches can be scaled to very large graphs, but are often sub-optimal.
In addition to these two families, learning approaches have been used to mix different heuristics \cite{Chung2019,Tian2020}.

\begin{figure}
    \centering
    \includegraphics[width=0.9\columnwidth]{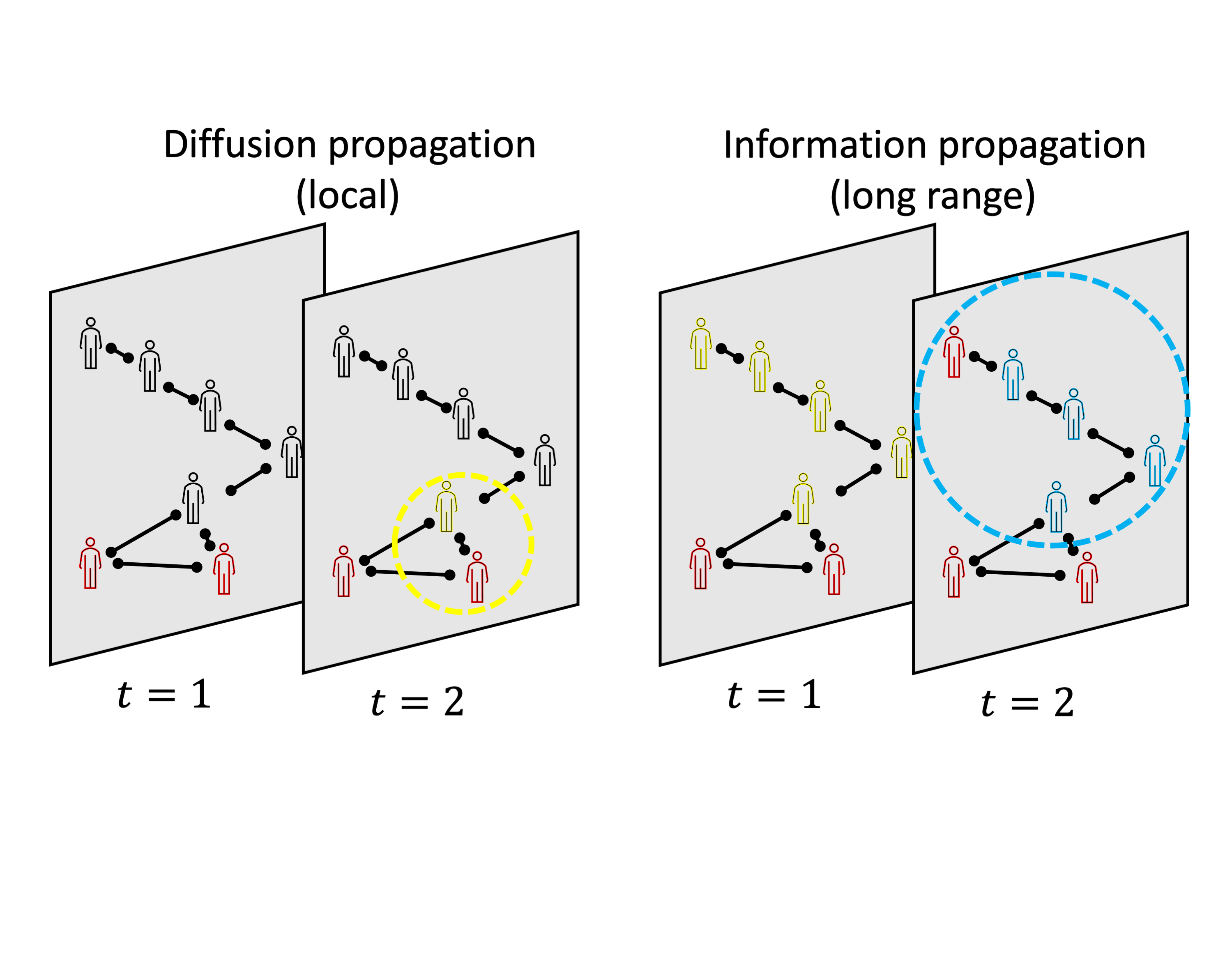}
    \caption{The difference between two types of data propagation on the graph. Red nodes are positively tested epidemic carriers. Yellow nodes are undetected epidemic carriers. Blue nodes are inferred to be infected. Left - infection propagation involves only direct neighbours. Right - long range information propagation: The top node is detected as infected at time $t=2$. As it must have been infected by its neighbor, our belief regarding the infection state of people on this long infection chain, including those that are found many hops away, change. We use two separate GNNs in order to model these two processes. }
    \label{fig:inf_vs_diff}
\end{figure}

We pose the problem of controlling a diffusive process on a temporally evolving graph as a partially-observed Markov decision process (POMDP). We then formulate the problem of selecting a subset of nodes for dynamical intervention as a {\em ranking } problem, and design an actor-critic RL algorithm to solve it.  We use 
the observed changes of nodes states and connections to construct a temporal multi-graph, which has time-stamped interactions over edges, and describe a deep architecture based on GNNs to process it.  
The main challenge in our setup is that the underlying dynamics is not directly and fully observed. Instead, partial information about the state of some nodes is given at each point in time. While the diffusive process spreads by point contacts, new node information may impact our belief on the state of a node a few hops away from the source of new information.
For example, consider an epidemic spreading on a network. Detecting an infected person directly modifies the probability that nodes that are connected to it by a path in the temporal graph are also infected (Fig.~\ref{fig:inf_vs_diff}). To address this issue, our architecture contains two separate GNN modules, one updates the node representation according to the dynamic process and the other is in charge of long range information propagation. These GNNs take as input a multi-graph over the nodes, where edges are time-stamped with the time of interactions. In addition, we show that combining RL with temporal graphs requires stabilizing information aggregation from other neighbors when updating nodes hidden states, and control how actions are sampled during training to ensure sufficient exploration. We show empirically the benefits of these components.

We test our approach on two very different problems, Influence Maximization and Epidemic Test Prioritization, and show that our approach outperforms state-of-the-art methods, often significantly. 
Our framework can be possibly further extended for problems beyond the ones mentioned here, e.g. traffic control, active  sensing for complex scenes, etc.



This paper makes the following  contributions:
    {\bf (1)}~A new RL framework for controlling \textit{partially-observed diffusive processes over graphs}. We present a novel formulation of two challenging problems: the \textit{testing allocation} problem and the \textit{partially-observed influence maximization} problem.
    {\bf (2)}~A new architecture for controlling the dynamics of diffusive processes over graphs. Our architecture prioritizes interventions on a temporal multi-graph by leveraging deep Graph Neural Networks (GNNs). 
    \textbf{(3)} A set of benchmarks and strong baselines, including network-based real-world contact tracing statistical data for COVID-19. Our RL approach achieves superior performance over these datasets.

\section{A motivating example}
\label{sec:toy-example}
\begin{figure}
\centering
  \includegraphics[width=0.6\columnwidth]{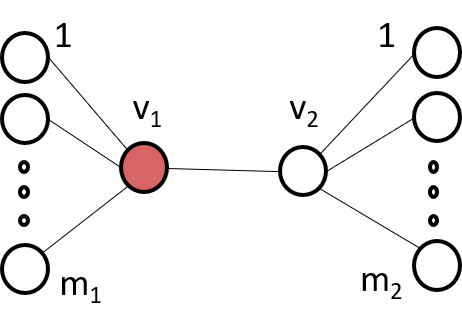}
    \caption{
    A double star configuration. The state of $v_{2}$ is unknown at the $t=1$. $v_1$ is infected at $t=0$.}
    \label{fig:toy-example}
    \vspace{-1mm}
\end{figure}

We begin with an example to illustrate the trade-offs of the problem (\figref{fig:toy-example}). In this example, our goal is to minimize the number of infected nodes in a social interactions graph.


Given a list of time-stamped interactions between nodes, we form a discrete time-varying graph as follows. 
If $u$ and $v$ interact at time $t$, then the edge $e\!=\!(u,v)$ exists at time $t$. Each interaction is characterized by a transmission probability $p_e(t)$, meaning that 
a healthy node that interacts with an infected node at time $t$ becomes infected with probability $p_e(t)$. 

For the purpose of this example, assume that we can test a single node only at odd timesteps. If the node is positively tested as infected, it is quarantined and cannot further interact with other nodes. Otherwise, we do not perturb the dynamics and it may interact freely with its neighbors. 

Consider the "two stars"  network in \figref{fig:toy-example}. The left hub (node $v_1$) has $m_1$ neighbors, and the right hub ($v_2$) has $m_2$.  At $t=0,$ only the edge $(v_{1},v_{2})$ is present with transmission probability $p$. For all $t\geq1$, all edges depicted in \figref{fig:toy-example} exist with transmission probability $1$. Assume that this is known to the agent, and that at $t=1$ we suspect that $v_{1}$ was infected at $t=0$. Clearly, we should either test $v_1$ or $v_2$. It is easy to compute the expected number of infected nodes in both cases (details in Appendix \ref{sec:appendix-toy-example}). The decision would be to test  $v_{2}$ if $2p\geq1+m_{1}/m_{2}$ and otherwise test $v_1$.


This example illustrates that an optimal policy must balance two factors: \emph{the probability that the dynamics is affected} - that a test action yields a ``positive", and the future consequences of our action - the \emph{strategic
importance} of selecting\emph{ $v_{1}$ vs. $v_{2}$}, expressed by the ratio $m_{1}/m_{2}$. 
A policy targeting likely-infected nodes will always pick node $v_1$, but since it only focuses on the first term and ignores the second term, it is clearly suboptimal.


\section{Problem Formulation} \label{sec:problem_formulation}

We start with a general formulation of the control problem, and then give two concrete examples from different domains: Epidemic test prioritization, and dynamic influence maximization. Formal definitions are given in Appendix \ref{sec:problem_formulation_appendix}.

\subsection{General formalism}
Consider a graph $G(t)=(\mathcal{V},\mathcal{E}(t))$
whose structure changes in time. $\mathcal{V}$ is the
set of nodes and $\mathcal{E}(t)=\{e_{uv}(t)\}$ is the set of edges
at step $t$.  Each edge $e_{uv}(t)$
is associated with features $\phi_{uv}(t)$ which may vary in time,
and each node $v$ is characterized with features $\zeta_{v}(t)$.

The state of a node $v\in\mathcal{V}$ is a random variable $ST_v(t)$ which can have values in  $\mathcal{Y}=\{y_{1},y_{2},..\}$. The node's state dynamic depends on the interactions between $v$ and its neighbors, its state and the state of those neighbors, all at time $t-1$.  At each step, the agent selects a subset $a(t)$ of $k$ nodes, and attempt to change the state of any selected node $v\in a(t)$, namely, apply a stochastic transformation on a subset of the nodes. Selecting nodes and setting their states defines the action for the agent, and plays the role of a knob for controlling the global dynamics of the process over the graph. 
The action space consists of all possible selections of  a subset $a(t)$ of $k$ nodes $a(t)\subset V$. Even for moderate graph, with $\sim100-1000$
and small $k$ the action space ${|\mathcal{V}| \choose k }$ is huge.



The optimization criterion depends only on the total number of nodes in  state $y_i$, $c_{i}(t)$.
The objective is therefore of the form $\max \sum_{t}\gamma^{t-t_{0}}g(c_{1}(t),c_{2}(t),..)$,
where future evaluations are weighted by a discount factor $\gamma \le 1$.
Additionally, the agent may be subject to constraints written
in a similar manner $\sum_{i}f_{i}(c_{1}(t),c_{2}(t),..)\geq z_{i}(t)$. 



\subsection{Epidemic test  prioritization}
We consider the recent COVID-19 outbreak that spreads through social contacts. The temporal graph $G$ is defined over a group of nodes (people) $\mathcal{V}$, and its edges $\mathcal{E}(t)$ are determined by their daily social interactions. An edge $(u,v)$ between two nodes exists at time
$t$ iff the two nodes interacted at time $t$. Each of these interactions is characterized by features $e_{uv}(t)$, including its duration, distancing and environment (e.g., indoors or outdoors). Additionally, each node $v$ has features $\zeta_{v}(t)$ (e.g., age, sex etc.).

\textbf{The SEIR model dynamics} \cite{Lopez2020}. Every node (person) can be in one of the following states: \textit{susceptible} -- a healthy, yet uninfected person ($S$ state), \textit{exposed/latent} -- infected but cannot infect others ($L$ state), \textit{infectious} -- may infect other nodes ($I$ state), or \textit{removed} -- self-quarantined and isolated from the graph ($R$ state). 

A healthy node can become infected by interacting with its neighbors. 
The testing intervention changes the state of a node. If infected or exposed, its state is set to $R$, otherwise it remains as it is.
 More details can be found in the appendix. 
 


\textbf{Optimization goal, action space.} The objective is to minimize the spread of the epidemic, namely, minimize the number of infected people (in either $L, R$ or $I$ states), over time. 
Our setup differs from previous work (e.g., \cite{Hoffmann2020, Wang2020}) in two important aspects. First, we do not assume a node can be vaccinated or immunized against the epidemic. Second, we do not assume a node can be quarantined or disconnected from the graph without justification, namely, without a positive test result. Often, nodes perform required social functionality.  Isolating a high-degree node from the network, like putting a bus-driver in quarantine, will either deteriorate the transportation network quality, or will require using a replacement driver that will have the same interactions pattern.  A preemptive node removal would either not affect the network connectivity or impair the network functionality.

\textbf{Observation space.} At each time $t$, the agent is exposed
to all past interactions between network nodes, $\left\{ \mathcal{E}(t')|t'<t\right\} $.
In addition, we are given partial information on the nodes state.
The agent is provided with information on a subset of
the infectious nodes at $t=0$. At every $t>0$, the agent observes all past test results, i.e, for every $v\in a(t'), t'<t$ we observe if  node $s$ was healthy at $t'$ or not. 

\subsection{Dynamic influence maximization}
The classical multi-round influence maximization problem \cite{domingos2001mining, Kempe2003, Lei2015} assumes the agent knows the groundtruth state of every node at every turn. More often than not, that is an unrealistic assumption. The agent can only know if a person is influenced if the person \emph{actively} signals it, for example by using a coupon code. Furthermore, there might be a substantial delay from the time the information was presented to the time a feedback was received. Therefore, we extend this setup to include partial observability.


\textbf{Model Dynamics.} Each node is either \emph{Influenced} or \emph{Susceptible}.  Influenced nodes try to influence their neighbors, following a dynamic generalization of two canonical models: Linear Threshold (LT) and Independent Cascades (IC).  In an IC model, if $u$ is Influenced and $(u,v)\in \mathcal{E}_t$, then $u$ may influence $v$ according to a probabilistic model. In a LT  model, each node $v$ is associated with a threshold $w_v$, and each edge $e$ carries an impact weight of $q_e$. If the sum of edge weights, the cumulative "peer pressure", of neighboring infected nodes exceeds $w_v$,  node $v$ is influenced. See Appendix \ref{sec:problem_formulation_appendix} for details on these models.


\textbf{Optimization goal, action space.} The goal is to maximize the number of \emph{Influenced} nodes. All nodes start at the \emph{Susceptible} state. At each step the agent selects a seed set $a(t)$ of $k$ nodes, and attempts to influence them. Each attempt succeeds with some probability $q$ independently for every $v\in a(t)$.


\textbf{Observation space.} At every step, an influenced node may reveal that it is influenced, e.g. by clicking on ads, with some probability $\eta$. The set of these signals at previous times along with past interactions between nodes consists the observation space.
\begin{figure}[t]
    \centering
    \includegraphics[width=0.85\columnwidth]{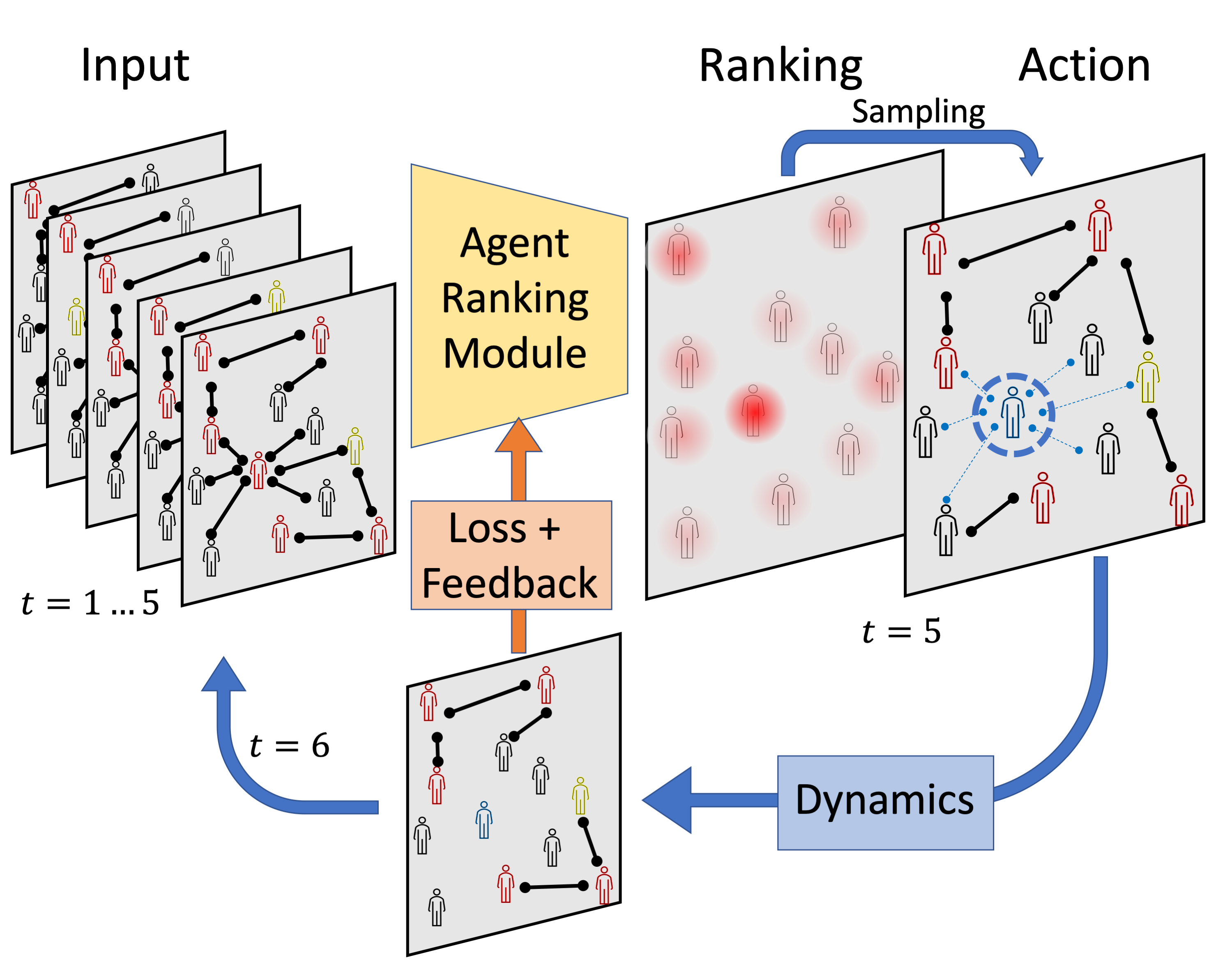}
    \caption{Schematic of our approach. The Ranking module receives as input a sequence of graphs and outputs scores over nodes. Scores are then used to sample actions, selecting nodes for intervention. 
    Here, the person circled in blue is selected for quarantine and its connections are canceled (dashed blue lines). The downstream effect on epidemic progression is then fed as a loss to the ranking module.}
    \label{fig:Approachsechematics}
\end{figure}

\section{Approach}
\label{sec:approach}

This section introduces our main contribution. Our goal is to select a subset of nodes for influencing the dynamics. The direct approach would be to perform a Monte Carlo simulation of the diffusive process for every possible action at every step, and choose the best performing action. However, this approach does not scale, and is unfeasible even for moderate networks \citep[see][and Appendix \ref{sec:approach-discussion_appendix} for discussion]{Liu2017, Banerjee2020}. An alternative popular approach uses predefined heuristics or greedy approaches (e.g., \cite{Yang2020,Preciado2014, Murata2018}), but this is arbitrary and often sub-optimal. 

We propose a learning-based approach, which generalizes from past patterns collected during training. Since our goal is to maximize an objective over time in a dynamic environment, RL is a natural choice (Figure \ref{fig:Approachsechematics}). 



Yet, even with a learning approach, solving the general case of the \textit{subset selection} problem would be combinatorially hard \cite{Kempe2003} and is difficult to scale to large graphs. At the other extreme, a simple approximated solution can be achieved by scoring each node independently and then selecting the top-ranked nodes. Unfortunately, this approximation would potentially be far from optimal because it neglects correlations across nodes that are crucial.   
Therefore, it is important that node selection would consider other nodes, at least locally. For example, creating tight clusters of \emph{Influenced} nodes is critical in Influence Maximization under the Linear Threshold model (see Appendix \ref{sec:problem_formulation_appendix}). Assume that the intervention budget  is sufficient for establishing a single cluster but there exist two equally beneficial regions to promote such cluster. The agent should learn to focus on one region rather than spread on two regions. This requires learning to choose optimal subsets rather than  choosing nodes independently.

Our approach takes a mid-road: We use a graph neural network to compute per-node scores, where each node is exposed to the features of nodes in its extended $m$-hop neighborhood (where $m$ is the depth of the GNN).  This way, agent can learn to take into account complex correlations, and to select high-quality subsets by ranking nodes by their scores.

\subsection{The Ranking Module} \label{sec:ranking_module} 


\paragraph{Overview.}
In our approach, an RL agent receives as input the node and edge features of the temporal graph, and scores each node. The module that performs that scoring is called the {\it ranking module} (\figref{fig:ranking}). Scores are used to generate a probability distribution over nodes, and then for sampling a subset of $k$ nodes for testing. Namely, the scores encode the agent policy.
The ranking module also updates the internal representation of each node, which aggregates past observations and information. 


The score of a node is affected both by propagation dynamics and by information available to the agent. One may hope that on a short time scale the node score would only be affected by its neighboring nodes. Unfortunately, 
information can propagate long distances in the graph almost instantaneously, because revealing the state of one node in a long chain affects other nodes.
 To handle this effect, the ranking module contains two GNNs (see Fig.~\ref{fig:ranking}). (1) A local diffusion component $D$ updates the diffusion process state; and (2) a long-range information component $I$ updates the information state. 

We use Proximal Policy Optimization (PPO) \cite{Schulman2017} to optimize our agent. We sequentially apply the suggested action, log the (state, action) tuple in an experience replay buffer, and train our model based on the PPO loss term.
We further motivate our framework and extended the discussion on our design choices in Appendix \ref{sec:approach-discussion_appendix}.


\subsection{Modules}
\begin{figure}[t]
    \centering
    \includegraphics[width=0.8\columnwidth]{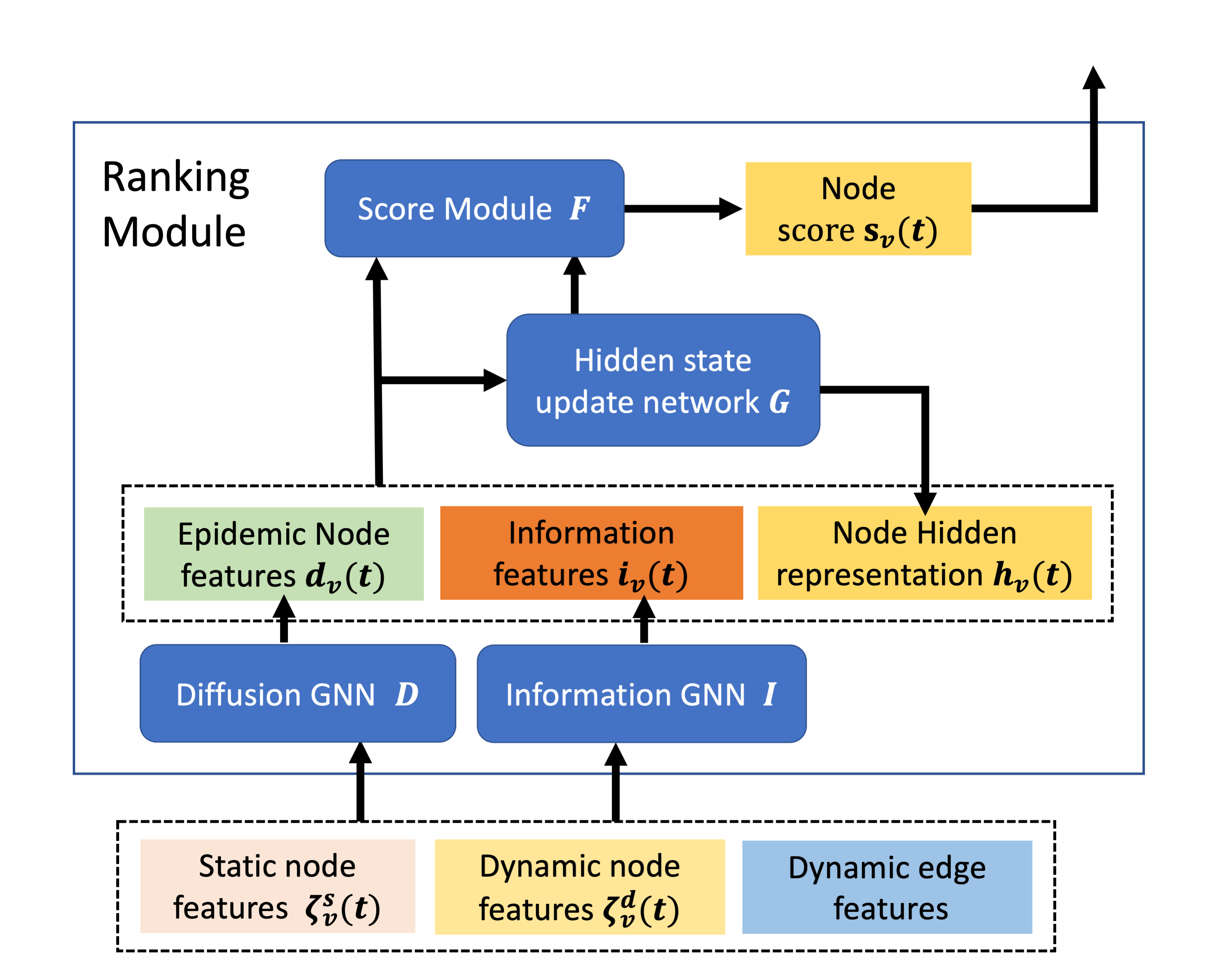}
    \caption{The ranking module. It is composed of 4 neural networks $I$,$D$,$G$,$F$, which update the nodes scores and hidden states at each time step.}
    \label{fig:ranking}
\end{figure}

\textbf{Input.} The input to the ranking module consists of three feature types:
(1) \textit{Static node features} $\zeta^s_v(t)$: e.g., topological graph centralities (betweenness, closeness, eigenvector, and degree centralities) and random node features.
(2) \textit{Dynamic node features} $\zeta^d_v(t)$ : All intervention results up to the current timestamp. We denote all nodes features as a concatenation $\zeta_v(t)=[\zeta^s_v(t),\zeta^d_v(t)]$.
(3) \textit{Edge features} and the structure of the temporal graph $\mathcal{E}(t)$: All previous interactions up to the current step, including the transmission probability for each interaction. All these  features  are  scalars, except the dynamic node features, which are encoded as one hot vectors. Figure \ref{fig:ranking} illustrates the basic data flow in the ranking module.

\textbf{Local diffusion GNN.} The spread through point contact is modeled by a GNN $D$. As the diffusive process spreads by only one hop per step, it is sufficient to model the spread with a single GNN layer. Formally, denote by $u\sim_{t}v$ an interaction between $u$ and $v$ at time $t$, and by $p_{vu}$ the probability of transmission during this interaction. For each $v$, the output of $D(\cdot)$ is a feature vector denoted by $d_v(t)$:
\begin{align*}
d_{v}(t) & =\sum_{u\sim_{t}v}p_{vu}(t)\cdot M_{e}(\zeta_{v}(t),\zeta_{u}(t);\theta_{m_e}), 
\end{align*}
where $M$ is multilayer perceptron (MLP).  Rather than considering the probability as an edge feature, this component mimics the dynamic process transition rule to accelerate learning.


\textbf{Long-range information GNN.}  GNN $I$ computes the {\em information state} of each node. As discussed above, updated information on a node $u$ a few hops away from node $v$ may abruptly change our belief on the state of $v$. Furthermore, this change may occur even if $v$ and $u$ did not interact in the last time step but rather a while ago. To update the information state, we construct a cumulative multi-graph $G'$ where the set of edges between nodes $v$ and $u$ at time $t$ are all the interactions that occurred during the last $\tau$ steps. 
The features of each edge  
$\phi_{vu}(t')$ at time $t'$ are the interaction delay $t-t'$ and the transmission probability $p_{v,v'}(t')$. The information features are the output of $k$-layer GNN; the $l^{th}$ layer is:
\[
x_{v}^{l}(t)=\sum_{v'\sim_{t}v}M^{l}(x_v^{l-1}(t),x_{v'}^{l-1}(t),\phi_{vv'}(t);\theta_M^l).
\]
As before, $M^l$ is an MLP, with $x_{v}^{0}(t)=\zeta_{v}(t)$ and $i_{v}(t)=x_{v}^{k}(t)$ are the final node features. 



\textbf{Score and hidden state update.} For every node we hold a hidden state $h_{v}(t)$, updated according to  a neural network $G$,
\begin{equation}
    h_{v}(t)=G(h_{v}(t-1),\zeta_{v}(t),d_{v}(t), i_{v}(t) ;\theta_{g})
\end{equation}
After updating the new node hidden states, we use them to calculate the node scores using a neural network $F$,
\begin{equation}
    s_{v}(t)= F(h_{v}(t),h_{v}(t-1),\zeta_{v}(t);\theta_{f})
\end{equation}

Here, $F$ and $G$ are two additional components (see Fig.~\ref{fig:ranking}). $F$ is an MLP, while $G$ can be either an MLP or recurrent module such as GRU.

\subsection{Sampling and scoring}
During inference, we pick the top $k$ scored nodes. During training, to encourage exploration we use the score per node $s_v(t)$ to sample $k$ nodes. 
We (1) map the score of $n$ nodes to a probability distribution 
(2) sample a node, and (3) adjust the distribution by removing its weight. We repeat this process $k$ iterations (sample without replacement).




\textbf{Score-to-probability.} Usually, node scores are converted to a distribution over actions using a softmax. As demonstrated in \cite{mei2020}, this approach is problematic as node probabilities decay exponentially with their scores, leading to two major drawbacks: it discourages exploration of low-score nodes, and also limits sensitivity to the top of the distribution rather than at the $k$-th ranked node. Instead, we set the probability to sample an action $a_{i}$ to
\begin{equation}
    \Pr(a_{i})=\frac{x_{i}'}{\sum x_{i}'}\quad \textrm{, with } x_{i}'=x_{i}-\min_{i}x_{i}+\epsilon,
    \label{eq: score-to-prob}
\end{equation}
where $\{x_{i}\}$ is the set of scores and $\epsilon$ a constant. 
The probability difference between low scoring nodes and high scoring nodes becomes less extreme than softmax. Furthermore, the parameter $\epsilon$ controls the initial exploration ratio. 
We compare our approach with the recent escort transform \cite{mei2020} that is considered to be a state-of-the-art score-to-probability method. As shown in Appendix \ref{sec:experimental details}, our method outperforms the escort transform in this problem.

\section{Experiments \label{sec: Experiments}}
We evaluated our approach in two tasks: (1) Epidemic test prioritization, and  (2) Dynamic influence maximization. More experiments and details are in Appendix \ref{sec:experimental details}.


\textbf{Real-World Datasets.} We tested our algorithm and baselines on graphs of different sizes and sources, ranging from 5K to over 100K nodes.
\textbf{(1) CA-GrQcA} A research collaboration network \cite{network_database}. \textbf{(2) Montreal}, based on WiFi hotspot tracing\cite{Hoen2015}. \textbf{(3) Portland:} a compartment-based synthetic network \cite{Wells2013,Eubank2004}. \textbf{(4) Email:} An email network \cite{Leskovec2007a} \textbf{(5) GEMSEC-RO: } \cite{rozemberczki2019gemsec}, friendship relations in the Deezer music service. 
All these networks have been extensively used in previous works, in particular in epidemiological studies, as key networks models \cite{Sambaturu, Yang2020, Herrera2016,Wells2013,Eubank2004}. Table \ref{tab:datasets-info}  summarizes the datasets. 

\textbf{Synthetic Datasets.} We considered three synthetic, random network families: \textbf{(1) Community-based networks} have nodes clustered into densely-connected communities, with sparse connections across communities. We use the \textit{Stochastic Block Model} (SBM, \cite{abbe2017community}), for 2 and 3 communities.
\textbf{(2) Preferential attachment (PA)} networks exhibit a node-degree distribution that follows a power-law (scale-free), like those found in many real-world networks.  We used the dual Barbarsi-Albert model \cite{Moshiri2018}.
\textbf{(3) Contact-tracing networks.}  We received anonymized high-level statistical information (see  Appendix \ref{sec:experimental details}) about real contact tracing networks, 
collected during April 2020.



\textbf{Generating temporal graphs.}  For all networks except CT graphs, at each  time step $t$ we select uniformly at random a subset of edges $\mathcal{E}(t)$ and then assign to each edge a transmission probability $q_e(t)$ sampled uniformly in $[0.5,1]$. We use a different methodology for the CT graphs,  See Appendix \ref{sec:experimental details} for details.

\textbf{Training procedure.} Algorithms were trained on randomly generated PA networks with $1000$ nodes. Each experiment was performed with at least three random seeds.

\subsection{Epidemic test prioritization}
\subsubsection{Baselines}
\label{sec:baselines}
We compare methods from three categories.



\textbf{A. Preprogrammed heuristic (no-learning) baselines.} Rank nodes based on: \emph{(1) Infected neighborhood}: Number of known infected nodes in their 2-hop neighborhood \cite{Meirom2015, 8071015}.
\emph{(2) Probabilistic risk}: Probability of infection at time $t-1$. Using dynamic programming to analytically solve the probability propagation. \emph{(3) Degree centrality} \cite{Salathe2010, Sambaturu}. \emph{(4) Eigenvector centrality:} \cite{Preciado2014,Yang2020}.

\textbf{B. Supervised learning.} Learn the risk per-node using features of the temporal graph, its connectivity, and infection state. 
Each time step $t$ and node $v_i$ is a sample, and its label is determined by the next step. 
%
\emph{(5) Supervised (vanilla).}
Features include a static component described in \secref{sec:ranking_module}, and a dynamic part that contains the number of infected neighbors and their neighbors. 
\emph{(6) Supervised (+GNN).} Like \#5, the input is the set of all historic interactions of $v_i$'s and its $d$-order neighbors.
\emph{(7) Supervised (+weighted degree).}
Like \#6, the loss weighs nodes are by their degree. 
%
\emph{(8) Supervised (+weighted degree +GNN).}
Like \#6 above, using degree-weighted loss like \#7.

\textbf{C. RL algorithms:}  \textbf{\RLGN{}} is our algorithm described in \secref{sec:approach}. The input to \textbf{(9) RL-vanilla} is the same as in (\#1) and (\#6) above. Correspondingly, the GNN module described in \secref{sec:approach} is replaced by a DNN similar to (\#6).

\textbf{Evaluation Metric.} The end goal of quarantining and epidemiological testing is to minimize the spread of the epidemic. Our success metric is therefore the percent of nodes kept healthy throughout the simulation. An auxiliary metric we sometime used was \textbf{\%contained:} The probability of containing the epidemic. This was computed as the fraction of simulations having cumulative infected nodes smaller than a fraction $\alpha=0.4$.

\subsubsection{Results}
\label{sec:results}

In the first set of experiments, we compared RLGN with the 9 baselines described in Section \ref{sec:baselines} on the synthetic networks described above. The results reported in Table \ref{tab:comaprison-chart-PA}
show that RLGN outperforms all baselines on all network types. 
We selected the top-performing algorithms and evaluated them on the large, real-world networks dataset. 

\begin{SCtable}
\centering
    \scriptsize
    \setlength{\tabcolsep}{5pt}
    \caption{ \% of healthy nodes achieved on a preferential attachment (PA) network, and contact tracing (CT) network. Here, two nodes were selected for testing at each step, $k=2$.} 
    \scalebox{0.99}{{\sc{
    \centering
        \begin{tabular}{lcc}
          & PA & CT\\
        \hline 
        Tree-based (2) &  $10\pm7$ & $11\pm3$\\
        Counter model (1) & $7\pm7$  & $14\pm5$ \\
        Degree  (3)   & $30\pm2$  &  $16\pm1$   \\
        Eigenvector (4)   & $30\pm1$  & $16\pm1$     \\
        SL (vanilla) (5) & $13\pm3$ & $17\pm1$\\
        SL + GNN (6) & $34\pm3$ & $32\pm$2\\
        SL + deg (7) & $15\pm3$ & $18\pm1$\\
        SL + deg + GNN (8)& $33\pm3$ & $32\pm1$\\
        \hline 
        RL (vanilla) (9) & $17\pm1$ & $16\pm1$\\
        RLGN (ours)  &   \textbf{52}{$\boldsymbol{\pm}$}\textbf{2} & \textbf{40}{$\boldsymbol{\pm}$}\textbf{1}  \\
        \hline 
        \end{tabular}
    }}}
\label{tab:comaprison-chart-PA}
\end{SCtable}



Table \ref{tab:dataset-healthy} compares the performance of the RLGN and the best baseline (SL) on the large-scale datasets. We included the centralities baselines (\#3,\#4) in the comparison as they are heavily used in epidemiological studies. Table \ref{tab:dataset-healthy} shows that RLGN consistently performs better than the baselines, and the gap is clearly statistically significant. 
We also evaluated the performance of RLGN on a Preferential Attachment network with $50,000$ nodes (mean degree $=2.8$), as this random network model is considered a reasonable approximation for many other real-world networks. 
The mean percentile of healthy nodes at the end of the episode was $51\pm1$ for RLGN, while for the SL+GNN it was only $21\pm2$, a difference of more than $15$ STDs.

\begin{figure}
    \centering
    \includegraphics[width=1.0\columnwidth]{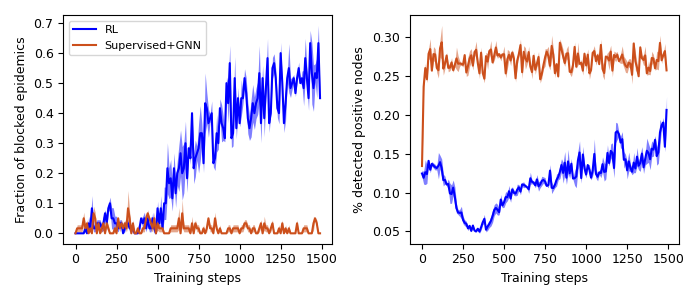}
    \caption{Supervised \textit{vs.} RL with 3-community networks. \textbf{Left:} RLGN successfully learns to contain the epidemic 60\% of the time (see containment definition in Appendix \ref{sec:experimental details}), while SL fails. \textbf{Right:} SL isolates many more infected nodes, but less important ones.}
    \label{fig:three_communities} 
\end{figure}

\textbf{Analysis.} To gain insight into these results, \figref{fig:three_communities} traces the fraction of contained epidemics and infected nodes during training in 3-community networks. Supervised learning detects \emph{substantially more infected nodes} than RLGN (right panel), but these tend to have a lower future impact on the spread, and it fails to contain the epidemic (left). A closer look shows that RLGN, but not SL, successfully learns to identify and neutralize the \emph{critical nodes that connect communities} and prevent the disease from spreading to another community. See a video highlighting these results  \href{https://youtu.be/Rhqy7YY9gX8}{online} \footnote{Link: \url{https://youtu.be/Rhqy7YY9gX8}}.

When would RLGN be successful? 
In sparsely connected networks, it is easy to cut long infection chains, and both approaches succeed. In  densely connected networks, there are no critical nodes, because there are many paths between any two nodes. This can also be viewed in terms of the $R_0$ coefficient, the mean number of nodes infected by a single diseased node. The greater $R_0$, the more difficult it is to contain the epidemic. Therefore, we expect RLGN to excel in intermediate regimes. 
Fig.~\ref{fig:stability}(a) indeed shows that RL has a significant advantage over supervised+GNN for a range of $R_0$ values between $2.0$ and $2.9$.

We have deepened our analysis, and investigated: (1) Can we quantify the algorithm by their ability to reduce $R_0$, the mean number of nodes infected by a single diseased node? Can we quantify the performance by the number of tests required to achieve the same level of performance, measuring their effective test utilization? (2) How robust the trained algorithms to variations in the epidemiological parameters? (3) How does the performance gap between the algorithms scale with the network size? Due to lack of space, we expand on these topics in Appendix \ref{sec:experimental details}. 

Appendix \ref{sec:experimental details} also includes a comparison between RLGN and the best performing baselines across a range of network sizes, initial infection sizes and testing capacities (Table \ref{tab:300-nodes-pa}).

\textbf{Ablation studies.} We assess the importance of key elements in our framework using ablation studies. First, to quantify the contribution of the information module, 
we removed it completely from our DNN module, keeping only the epidemic module. The full DNN module achieved a contained epidemic score of $0.77\pm0.06$, while the ablated DNN module corresponding score was $0.62\pm0.10$, a degradation of more than $20\%$. This shows that the information module has a critical role in improving the performance of the RLGN framework.

Second, in the opposite direction, one may wonder: Why separate local and long-range GNNs, rather than a single higher-capacity network? We found that using a local-diffusion GNN training converges faster (Fig. \ref{fig:local_diffusion}). Presumably, because it models the process more closely to the true spreading.

Appendix \ref{sec:experimental details} contains additional ablation studies of key elements in our framework:
(A) Our score-to-probability function outperforms the popular softmax distribution and escort transform. (B) Internal state normalization in scale-free networks accelerates training substantially.

\ignore{
\begin{table*}
\centering{}%
\caption{\label{tab:dataset-healthy}The mean percentile of healthy nodes after
a 20 steps. RLGN performs better on all datasets with a statistically
significant gap. At each step 1\% of the nodes are tested.}
\begin{tabular}{|l|c|c|c|c|c|}
\hline 
 & CA-GrQc & Montreal & Portland & Email & GEMSEC-RO\\
\hline 
\hline 
Degree & $25.52\pm0.01$  & $12.83\pm0.01$ & $0.67\pm0.01$ & $71.14\pm0.02$ & $2.43\pm0.01$\\
\hline 
E.vector & $25.37\pm0.01$ & $8.06\pm0.01$ & $0.04\pm0.01$  & $55.10\pm0.02$ & $2.41\pm0.01$\\
\hline 
SL & $29.79\pm0.02$ & $23.09\pm0.03$ & $1.57\pm0.01$ & $68.45\pm0.05$ & $4.26\pm0.01$\\
\hline 
RLGN & \textbf{42.69}{$\boldsymbol{\pm}$}\textbf{0.03}  & \textbf{39.68}{$\boldsymbol{\pm}$}\textbf{0.03} & \textbf{3.71}{$\boldsymbol{\pm}$}\textbf{0.01} & \textbf{89.19}{$\boldsymbol{\pm}$}\textbf{0.02}  & \textbf{6.52}{$\boldsymbol{\pm}$}\textbf{0.01} \\
\hline 
\end{tabular}
\end{table*}
}

\setlength{\tabcolsep}{2pt}
\begin{table}[t]
\centering{}%
    \small
    \caption{\label{tab:dataset-healthy}Mean percentile of healthy nodes after 20 steps. RLGN perform better on all datasets. In all cases, std $<0.1$. 1\% of nodes are tested at each step. }
    \begin{tabular}{lccccc}
    \toprule
     & CA-GrQc & Montreal & Portland & Enron & GEMSEC-RO\\
    \midrule
    Degree & $25.5$  & $12.8$ & $0.7$ & $71.1$ & $2.4$\\
    E.vector & $25.4$ & $8.1$ & $0.04$  & $55.1$ & $2.4$\\
    SL & $29.8$ & $23.1$ & $1.6$ & $68.5$ & $4.3$\\
    RLGN  & \textbf{42.7}  & \textbf{39.7}& \textbf{3.71} & \textbf{89.2}  & \textbf{6.5} \\
    \bottomrule 
    \end{tabular}
\end{table}




\subsection{Influence Maximization}

\setlength{\tabcolsep}{4pt}
\begin{table*}
\centering{}%
\vskip 0.15in
    \small
\caption{Influence Maximization: Mean percentile of influenced nodes after 15 steps \label{tab:IM-geometric}.}
\centering
    \begin{tabular}{lccccc}
    \toprule
    & CA-GrQc & Montreal & Enron & GEMSEC-RO & CA-HEPTh \\
    \midrule 
    LIR \cite{Liu2017} & $7.3\pm0.3$ & $86.2\pm0.7$ & $29\pm0.3$ & $0.25\pm0.02$ & \textbf{$\boldsymbol{9.2\pm0.3}$}\\
    LIR (filtered) & $8.0\pm0.2$ & $\boldsymbol{86.4\pm0.7}$ & $28.8\pm0.3$ & $0.22\pm0.02$ & $8.5\pm0.3$\\
    Degree & $8.4\pm0.2$ & $85.5\pm0.8$ & $\boldsymbol{31.6\pm0.6}$ & $0.07\pm0.01$ & \textbf{$\boldsymbol{9.2\pm0.3}$}\\
    Degree Discounted \cite{Murata2018} & $8.7\pm0.2$ & $85.6\pm0.7$ & $26.7\pm0.6$ & $0.05\pm0.01$ & $8.4\pm0.2$\\
    Eigenvector \cite{Bozorgi2016} & $8.3\pm0.2$ & $82.9\pm0.8$ & $\boldsymbol{31.8\pm0.5}$ & $0.07\pm0.01$ & $2.2\pm0.2$\\
    RLGN (ours)& $\boldsymbol{10.2\pm0.6}$ & $\boldsymbol{87.4\pm0.5}$ & \textbf{$\boldsymbol{31.3\pm0.6}$} & $\boldsymbol{5.8\pm0.3}$ & \textbf{$\boldsymbol{9.1\pm0.5}$}\\
    \bottomrule
    \end{tabular}

\end{table*}

\textbf{Baselines}. Unlike the epidemic test prioritization, in this problem there is no supervised signal; there is no immediate feedback that may be used for supervision. We compare our RLGN framework against the state-of-the-art scalable algorithms. \emph{(1) LIR} \cite{Liu2017} is an algorithm for top-k ranking for the IM problem. It was shown to achieve similar performance to MC based methods. \emph{(2) LIR (filtered):} LIR was designed for a fully observable setup. We extend this algorithm to a partially observed setup and filter out nodes with an identified \emph{influenced} neighbor. The motivation is that it is likely that such nodes are already influenced or likely to be influenced soon. \emph{(3) Degree discounted} \cite{Chen2009} is a topology-based algorithm that was shown to achieve a state-of-the-art performance on some networks, and \emph was recently extended to temporal graphs \cite{Murata2018}. \emph{(4) Degree Centrality} and \emph{(5) Eigenvector Centrality}, defined previously, were also used extensively \cite{Lei2015,Chen2014,Bozorgi2016}.

\textbf{Results}. We have compared RLGN against the aforementioned baselines on the real-world datasets in Table \ref{tab:IM-geometric}. We included an additional (CA-HEPTh) network that was frequently used as a benchmark for this problem.

 Table \ref{tab:IM-geometric} shows that RLGN performs remarkably in this domain as well. It achieves state-of-the-art performance, often with a considerable gap. Additional experiments and experimental details appear in Appendix \ref{sec:experimental details}.

\section{Previous work \label{sec:related_work}}

\textbf{Deep Learning on graphs.} Graph neural networks (GNNs) are deep neural networks that can process graph-structured data \cite{Sperduti1993, Sperduti1994, Sperduti1997, Pollack1990, Kuchler1996, kipf,Gilmer2017,Duvenaud2015, Hamilton2017, Velickovic2017}. Several works combine recurrent mechanisms with GNNs to learn temporal graph data, \cite{liu2019towards, rossi2020temporal, liu2020towards, pareja2020evolvegcn}. Further information can be found in  \cite{kazemi2020representation}.

\textbf{Ranking on graphs.} The problem of ranking on graphs is a fundamental CS problem. It has various applications such as web page ranking \cite{page1999pagerank,agarwal2006ranking} and knowledge graph search \cite{xiong2017explicit}.

\textbf{Reinforcement learning and graphs} studies can be split into two main categories: leveraging graph structure for general RL problems \cite{Zhang2018,Jiang2018}, and applying RL methods for graph problems. Our work falls into the latter. An important line of work uses RL to solve NP-hard combinatorial optimization problems on graphs. \cite{Zhu,Dai, Wei2021}. 

\textbf{Manipulation of dynamic processes on graphs. } 
The problem of node manipulation (e.g., vaccination) for controlling epidemic processes on graphs was intensively studied \cite{Hoffmann2020, Medlock2009}. This problem is often addressed in the setup of the fire-fighter problem and its extensions \cite{firefighter_survey, Tennenholtz2017, Sambaturu}. Other work considered the problem of vaccination assignments, and cast this problem into a minimal cover problem \cite{Wang2020,Song2020, Wijayanto2019}. Other common approaches include developing centrality measures designed to highlight bottleneck nodes \cite{Yang2020}, or using spectral methods for allocating resources \cite{Saha2015, Preciado2014, Ogura2017}. Alternative line of research \cite{MILLER2007780, Cohen2002} developed heuristics for the same task.

In most previous work setups a single decision is taken. In our multi-round setup, the agent performs a sequential decision making. The agent needs to balance between retrieving information (for better informed future decisions), maximizing the probability that the intervention will be successful, and optimizing the long-term goal. 

 

\textbf{Influence Maximization}, (IM) is a canonical optimization problem of dynamical processes on graphs. IM was first presented in \cite{Kempe2003}, and proved to be NP-Hard and hard to approximate. Key approximation algorithms were derived in \cite{Goyal2011, Nguyen2016}, but since they do not scale to large graphs, many alternative heuristics were developed \cite{Murata2018,Liu2017}. For  surveys, see \citet{Banerjee2020, Li2018}. Multi-armed Bandit was used for estimating model parameters  \cite{Vaswani2017, Lei2015}. 
The IM formulation was extended to a multi-round framework by \citet{Lin2015}. \citet{Chung2019, Tian2020, Lin2015} used RL to find the optimal combination of heuristics from a short list of  hand-designed features.

These approaches are limited by the small number of pre-selected heuristics and by the problem-specific, hand-crafted features. In contrast, our approaches is general and it is not limited to reweighting or a predefined subset of policies, neither uses hand-designed, problem-specific features. Our agent learns a policy from scratch and uses GNNs to generalize to different domains \cite{Yehudai2020}. 

\section{Conclusions}
This paper shows that combining RL with GNNs provides a powerful approach for controlling diffusive processes on graphs.
Our approach handles an exponential state space, combinatorial action space and partial observability, and achieves superior performance on challenging tasks on large, real-world networks.

The approach and model discussed in this paper can be applied to important problems other than epidemic control and influence maximization. For example, fake news can be maliciously distributed, and spread over the network. A decision maker can verify the authenticity of items, but only verify a limited number of items per a time period. The objective would be to minimize the total number of nodes that observe fake items.

Our approach assumes that a decision taken by considering only $k$ hops neighborhood of each node is a fairly good approximation to the optimal policy which takes into account the whole graph. If long range correlations exists, this may deteriorate performance. As such, it is sufficient to train our model on small graphs and infer on a larger graph. An interesting question is the ability of our approach to address edge cases that may result from this training protocol, and the generalization ability of our model as a function of long-range correlations in the data.

A key concern for real world application is privacy preservation of individual nodes. Our approach requires local aggregated information about the node's neighborhood, compared to other approaches (e.g., \cite{Kempe2003,Yang2020} which required detailed information on the complete graph. Furthermore, recent papers \cite{zhou2021vertically} have shown that it is possible to use graph neural network while preserving privacy, and we leave it for future research to apply such approaches in this setup.

\bibliographystyle{icml2021}
\bibliography{ref,corona_paper}
\onecolumn
\appendix
\icmltitle{Controlling Graph Dynamics with \\ Reinforcement Learning and Graph Neural Networks : \\ Supplementary Material}

\setcounter{table}{0}
\renewcommand{\thetable}{S\arabic{table}}
\setcounter{figure}{0}
\renewcommand{\thefigure}{S\arabic{figure}}
\section{Motivating example details\label{sec:appendix-toy-example}}

We provide the details for the example from \secref{sec:toy-example} (\figref{fig:toy-example}). Recall that Our goal is to minimize the number of infected nodes in a social interactions graph.
A natural algorithmic choice would be to act upon nodes that are  most likely infected. The following example shows why this approach is suboptimal.

We form a time-varying graph from a list of interactions between nodes at various times. If $u,v$ interacted at time $t$ then the edge $e=(u,v)$ exists at time $t$. Each interaction is characterized by a transmission probability $p_e(t)$.  If a node was infected at time $t$ and its neighbor was healthy, then the healthy node is infected with probability $p_e(t)$. 
Assume that we can test a single node at odd timesteps. If the node is identified as infected, it is sent to quarantine and cannot further interact with other nodes. Otherwise, we do not perturb the dynamics and it may interact freely with its neighbors.

Consider the "two stars"  network in \figref{fig:toy-example}. The left hub (node $v_1$) has $m_1$ neighbors, and $m_2$ nodes are attached to the right hub $v_2$, with $m2\gg m1$.  At $t=0,$ only the edge $e=(v_{1},v_{2})$ is present with $p_e(t=0)=p$. Then, for all $t\geq1$, all edges depicted in \figref{fig:toy-example} exist with $p_e(t)=1$. Assume that this information is known to the agent, and that at $t=1$ we suspect that node $v_{1}$ was infected at $t=0$. 

If $v_1$ would turn out to be healthy, than any test would result in a negative result and would not affect the dynamics. Hence, in the following derivation we condition on node $v_1$ being infected at $t=0$. Note that we can not quarantine prematurely $v_1$ unless detected as positive. In this case, we clearly should test either $v_{1}$ or $v_{2}$. We can compute the expected cost of each option exactly.
\textbf{Alternative I:} Test ${v_2}$. With probability $p$, $v_{2}$ becomes infected at $t=1$, and we block the epidemic from spreading. However, we forfeit protecting $v_{1}$ neighbors, as all of them will be infected in the next step. With probability $1\!-\!p$ the test is negative, and we fail to affect the dynamics. At $t=2$ node $v_{2}$ will get infected and at $t=3$ all of $v_{2}$'s neighbors become infected too, ending up with a total of $\left(m_{2}+1\right)$ infections. The expected cost in choosing to test $v_{2}$ is $(1-p)\cdot m_{2}+m_{1}$.  
\textbf{Alternative II:} Test $v_{1}$. We block the spread to $v_{1}$'s neighbors, but sacrifice all $m_{2}$ neighbors of $v_{2}$ with probability $p$. The expected cost in choosing $v_2$ is $p\cdot m_{2}$. 
The decision would therefore be to test for $v_{2}$ if $2p\geq1+m_{1}/m_{2}$. 

This example illustrates that an optimal policy must balance two factors: \emph{the probability that the dynamics is affected} - that a test action on $v_2$  yields a ``positive", measured by $p$, and the future consequences of our action - the \emph{strategic
importance} of selecting\emph{ $v_{1}$ vs. $v_{2}$}, expressed by the ratio $m_{1}/m_{2}$. 
A policy targeting likely-infected nodes will always pick node $v_1$, but since it only focuses on the first term and ignores the second term, it is clearly suboptimal.

\section{Problem Formulation - Additional Details} \label{sec:problem_formulation_appendix}

In this section we fill in on the details of our setup.

\subsection{Epidemic test  prioritization}
\textbf{The SEIR model dynamics} \cite{Lopez2020}. Every node (person) can be in one of the following states: \textit{susceptible} -- a healthy, yet uninfected person ($S$ state), \textit{exposed/latent} -- infected but cannot infect others ($L$ state), \textit{infectious} -- may infect other nodes ($I$ state), or \textit{removed} -- self-quarantined and isolated from the graph ($R$ state). Formally, let $\mathcal{I}(t)\subset V$ be the set of infectious nodes at time $t$, and similarly $\mathcal{L}(t)$, $\mathcal{R}(t)$
and  $\mathcal{S}(t)$ be the sets of latent(exposed), removed and susceptible (healthy) nodes. 

A healthy node can become infected by interacting with its neighbors. 
Each active edge at time $t$, $e\in\mathcal{E}(t)$, carries a transmission probability $p_{e}(t)$. Denote the set of impinging
edges on node $v$ with an infectious counterpart at time $t$ by $E_{v}(t)$. Formally,
$$E_{v}(t)=\left\{ e\in\mathcal{E}(t)|e=(v,u),SV_u(t-1)=I\right\} .$$
The probability of a healthy node to remain healthy at time $t$ is $$\prod_{e\in E_{v}(t)}\left(1-p_{e}(t)\right)$$
otherwise it becomes infected, but still in a latent state. Denote the time of infection of node $v$ as $T_{v}.$ A node in a latent state will stay in this state at time $t$ if $t<T_{v}+D_{v}$, where $D_{v}$ is a random variable representing the latency period length, otherwise its state changes to infectious. 
The testing intervention changes the state of a node. If infected or exposed, its state is set to $R$, otherwise it remains as it is. In principle, a node is state $R$ can be restored to the network after quarantining, though in our setup the quarantining period is larger than the episode duration and therefore once a node is in $R$ it remains detached from the network for the rest of the simulation.

\textbf{Optimization goal, action space.} The objective is to minimize the spread of the epidemic, namely, minimize the number of infected people (in either $L, R$ or $I$ states), 

$$
\min\sum_{t,v}\gamma^{t}\left\lVert \mathcal{L}(t) \cup \mathcal{R}(t) \right\rVert,
$$
where $\gamma\in(0,1]$ is a discount factor representing the relative importance of the future compared to the present. We used $\gamma=0.99$ throughout the paper.

The action space consists of all possible selections of  a subset $a(t)$ of $k$ nodes $a(t)\subset V$. Even for a moderate graph, with $\sim100-1000$
and small $k$ the action space ${|\mathcal{V}| \choose k }$ is huge.

\subsection{Dynamic influence maximization}

\textbf{Model Dynamics.} Each node is either \emph{Influenced}, denoted by $I$ or \emph{Susceptible} ($S$).  At each time the agent selects a seed set $a(t)$ of $k$ nodes, and attempt to influence them to its cause. This succeeds with probability $q$ independently for every $v\in a(t)$. Influenced nodes then propagate this cause, following a dynamic generalization of two canonical models: Linear Threshold (LT) and Independent Cascades (IC).

In a Linear Threshold dynamic model, each node $v$ is associated with a threshold $w_v$, and each edge $e$ carries an impact weight of $q_e$. The "peer pressure" $z_v(t)$ on a node is the total weight of active edges in the last $T_{peer}$ steps connecting influenced neighboring nodes and node $v$. 
$$z_v(t) = \sum_{e\in E_v(t)} q_e, \quad E_v(t) = \left\{(u,v)| (u,v)\in \mathcal{E}(t'),t-t'<T_{peer}, ST_u(t)=I \right\}$$

If the "peer pressure" on node $v$ exceeds $w_v$, node $v$ state is changed to \emph{Influenced}. 

In an Independent Cascades model, if $u$ is Influenced and $(u,v)\in \mathcal{E}_t$, then $u$ attempts to influence $v$. We explored two variations, IC(constant) and IC(geometric). In IC(constant), the success probability of each attempt is fixed as some $p$, while in IC(geometric), the success probability decays with the number of influence attempts $m_{(u,v)}$,  so the success probability is $p^{m_{(u,v)}}$. This mimics the reduced effect of presenting the same information multiple times.

\section{Approach Discussion}
\label{sec:approach-discussion_appendix}

In this section we further motivate our design.

\textbf{Policy gradients.} An action-value approach like Q-learning implies that an approximate value is assigned to every possible action. The action space of choosing a subset of $k$ nodes out of $n$ nodes is prohibitively too large even for small $n$ and $k$. Instead, we use a policy-gradient algorithm and model the problem as a ranking problem.


Many on-policy gradient algorithms use entropy to define a trust region.  Computing the entropy requires summing ${ |\mathcal{V}| \choose k }$ terms at each step, and it is computationally expensive. A more scalable solution is the unbiased entropy estimator of \cite{Zhang2018a}, but the variance of that estimator is high. As an alternative, 
PPO trust region is not based on an explicit evaluation of the entropy, and performed better in our experiments. We also evaluated A2C, which did not perform as well as PPO in our experiments.

\textbf{Critic module.} PPO, as an actor-critic algorithm, requires a critic module to estimate the value function in a given state. We construct the critic using an architecture similar to the ranking module. We apply an element-wise max operation on the rows (representing the nodes) of the score module $F$'s input (\figref{fig:ranking}). This reduces $F$'s input to a single row of features, and the output is then a scalar rather than a vector. Importantly, the critic is parametrized by a different set of weights than the ranking module (actor).

\textbf{Normalization in scale-free networks.} Recurrent neural networks are well-known to suffer from the problem of exploding or vanishing gradients. This problem is exacerbated in a RNN-GNN framweork. A node in Graph Neural Networks framework receives updates from a large number of neighbors and its internal state may increases in magnitude. The next time that the GNN module is applied (e.g., at the next RL step), the node's updates its neighbors, and its growing internal state updates and increases the magnitude of the internal state of its neighbors.  This leads to a positive-feedback loop that causes the internal state representation to diverge. This problem is particularly severe if the underlying graph contains hubs (highly connected nodes). Scale-free networks contain with high probability "hub" nodes that have high-degree, namely $O(n)$ neighbors. The presence of these hubs further aggravates this problem.  Since RL algorithms may be applied for arbitrary long periods, the internal state may grow unbounded unless corrected.

One approach to alleviate this problem is by including an RNN like a GRU module, where the hidden state values pass through a sigmoid layer. As the magnitude of the input grows, the gradient become smaller and training slows down. Alternatively, This problem can be solved by directly normalizing each node hidden state. We have experimented with various normalization methods, and found that $L_2$ normalization worked best, as shown in the next section.

\textbf{Transition probabilities}. In the case of the COVID-19 pandemic, the transition probabilities can be estimated using the interaction properties, such as duration and inter-personal distance, using known epidemiological models. This was done by the government agency which provided our contact tracing network (see below). Alternatively, one can learn the transmission probability as a regression problem from known interactions (e.g, using data from post-infection questioning). Finally, if this information is not accessible, it is possible to omit the epidemic model \emph{E} from the proposed framework and use only the feature vector created by the information module \emph{I}.

In a scale free network, there exists hubs with $O(n)$ neighbors. 
As a simple case, consider a star graph with a large number of nodes. In a GNN framework, it receives updates from a large number of neighbors, and its internal state increases in magnitude. In the next application of the GNN module, e.g., in the next RL step, its growing internal state will induce an increase in the magnitude of its neighbor's internal state, resulting in a positive feedback loop that will blow the internal state representation. Fundamentally, an RL algorithm may be applied for arbitrary long episodes, which will allow the internal state to grow unbounded. 

\section{Additional Experimental details \label{sec:experimental details}}
In this appendix we expand of various experimental aspects. We first elaborate on the different baselines.

\begin{figure}
    (a)\hspace{120pt}
    (b)\hspace{120pt}\\
    \includegraphics[width=0.48\columnwidth]{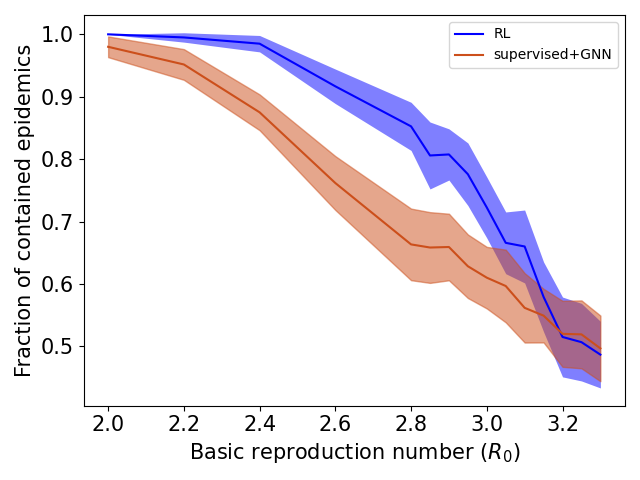}
    \includegraphics[width=0.48\columnwidth]{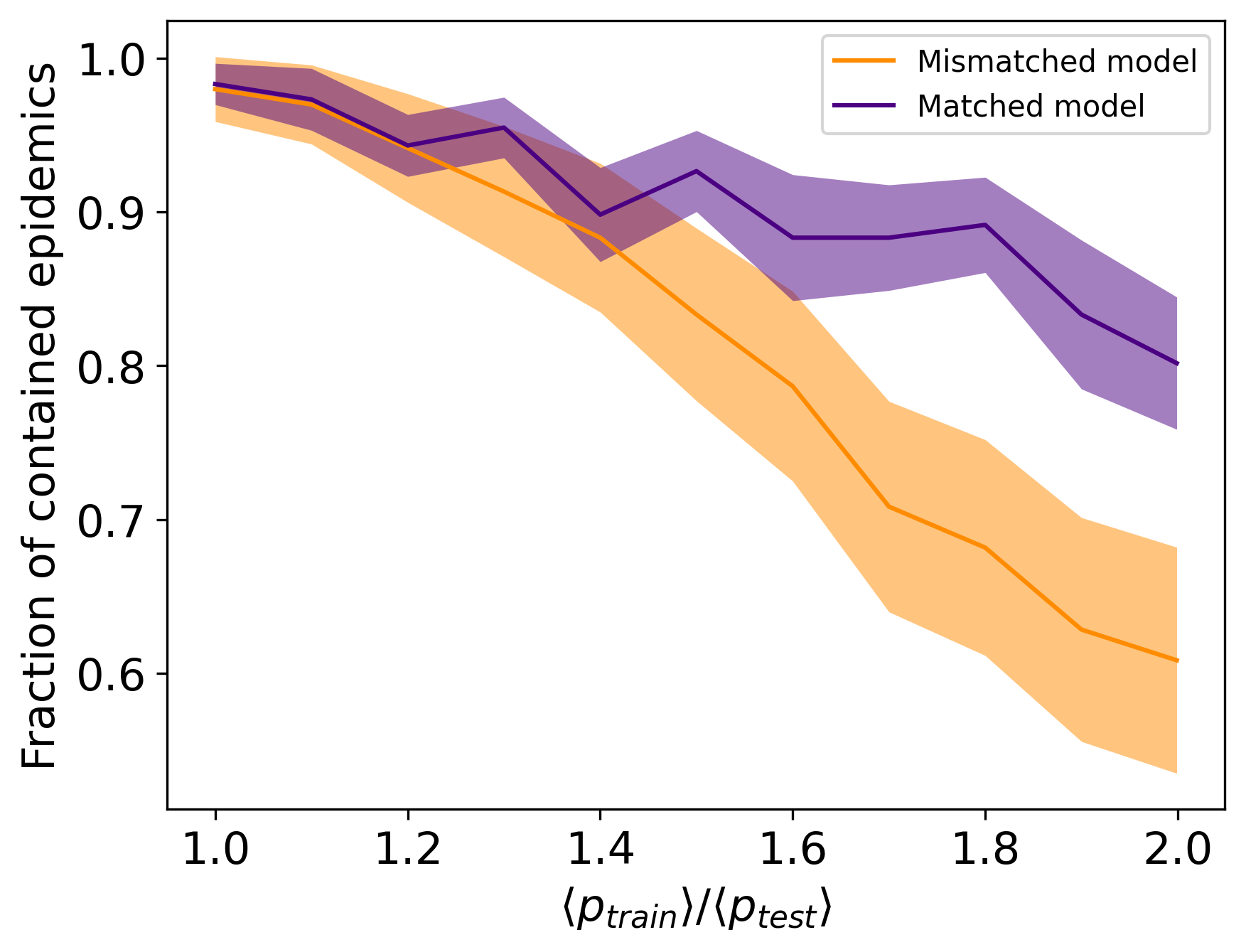}
    \caption{\textbf{Stability analysis:} \textbf{(a)} The contained epidemic fraction as a function of the basic reproduction number $R_0$ on a PA network. RLGN outperforms SL over a large range of $R_0$ values. \textbf{(b)} Stability against test-time shift in transmission probability. \textit{Orange:} The performance of RLGN deteriorates when the mean transmission probability at test time is higher more than $40\%$ than train time. \textit{Purple:} As a baseline, training and testing with the same higher transmission probability.  
    \label{fig:stability}
    }
\end{figure}

\subsection{Synthetic datasets} 
We study three types of networks which differ by their connectivity patterns.


\textbf{(1) Community-based networks} have nodes clustered into densely-connected communities, with sparse connections across communities. We use the \textit{Stochastic Block Model} (SBM, \cite{abbe2017community}), for 2 and 3 communities.
The Stochastic Block Model (SBM) is defined by (1) A partition of nodes to $m$ disjoint communities $C_{i}$, $i=1\dots m$; and (2) a matrix $P$ of size $m\times m$, which represents the edge probabilities between nodes in different communities, namely, the matrix entry $P_{i,j}$ determines the probability of an edge $(v,v')$ between $v\in C_{i}$ and $v'\in C_{j}$. The diagonal elements in $P$ are often much larger than the off-diagonal elements, representing the dense connectivity in a community, compared to the intra-connectivity between communities.

%

\textbf{(2) Preferential attachment (PA)} networks exhibit a node-degree distribution that follows a power-law (scale-free), like those found in many real-world networks.  We use the dual Barbarsi-Albert model \cite{Moshiri2018}, an extension to the popular Barabasi-Albert model \cite{Barabasi1999}, 
which allows for continuously varying the mean node degree. The node degree of the resulting network has a power-law distribution. 



\textbf{(3) Contact-tracing networks.} 
We received anonymized high-level statistical information about real contact tracing networks that included the distribution of node degree, transmission probability and mean number of interactions per day, collected during April 2020.

\begin{figure}
    \centering
   \begin{tabular}{cc}
       \includegraphics[width=0.5\columnwidth]{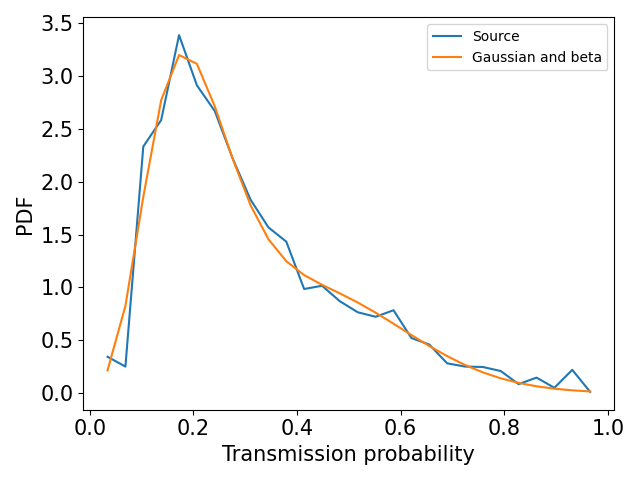}& \includegraphics[width=0.5\columnwidth]{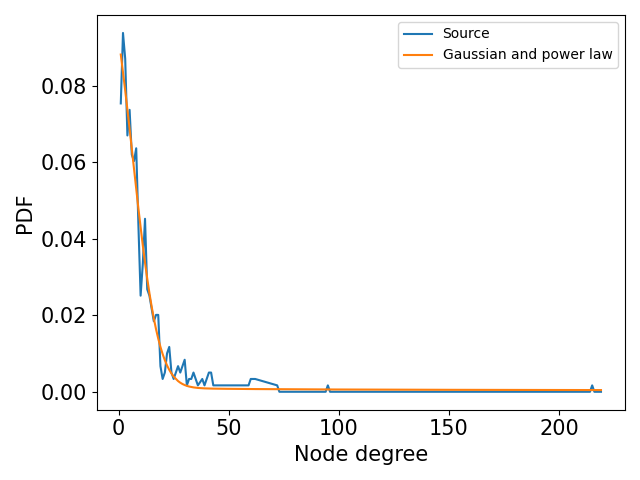}\\
       (a) & (b)
       \end{tabular}
       \caption{Statistics of a real-world contact-tracing graph. (a) The empirical transition probability $P(p_e)$ on a contact tracing network and our suggested curve fit. (b) The degree distribution on the contact tracing network, along with its fit.}
       \label{fig:transmission_prob}
\end{figure}

Fig. \ref{fig:transmission_prob}(b) presents the degree distribution in this data, and the transmission probability is presented in Fig. \ref{fig:transmission_prob}(a). The latter was derived based on the contact properties, such as the length
and the proximity of the interaction. On average, $1.635\pm0.211$ interactions with a significant transmission probability were recorded per-person per-day. We generated random networks based on these distributions using a configuration model framework \cite{Newman2010a}. The fitted model for the degree distribution is a mixture of a Gaussian and a power-law distribution
\begin{align}
    P(degree=x)=2.68\cdot\mathcal{N}(-4.44,11.18)+3.2\cdot10^{-3}\cdot x^{-0.36}.    
\end{align}
The fitted model for the transmission probability is a mixture of a Gaussian and a Beta distribution
\begin{align}
    P(p_{e}=x)=0.47\cdot\mathcal{N}(0.41,0.036)+0.53\cdot Beta(5.05,20.02).
\end{align}

\textbf{Evaluation on CT data.} 
Due to privacy constraints, we did not have access to ``live" CT graphs. We used these statistics to generate topologically similar synthetic graphs. Likewise, we simulated activity patterns based on activity statistics of the real CT network.

\subsection{Epidemic test prioritization baselines}

\textbf{A. Preprogrammed Heuristics.} 
The most prevalent baseline, used in practice nowadays in a few countries and circumstances, is based on the proximity of a node to infectious node.
We compare with two such methods to ran
k nodes. 
\textbf{(1) Infected neighbors.} Rank nodes based on the number of known infected nodes in their 2-hop neighborhood (neighbors and their neighbors). Each node $v$ is assigned a tuple $(I_{v}^{(1)},I_{v}^{(2)})$, and tuples are sorted in a decreasing lexicographical order. A similar algorithm was used in \cite{Meirom2015, 8071015} to detect infected nodes in a noisy environment, and its error was shown to vanish  asymptotically in the network size. 
\textbf{(2) Probabilistic risk.} Each node keeps an estimate of the probability it is infected at time $t-1$. To estimate infection probability at time $t$, beliefs are propagated from neighbors, and dynamic programming is used to analytically solve the probability update. See Appendix \ref{sec:tree model} for details. \textbf{(3) Degree centrality.} In this baseline high degree nodes are prioritized. This is an intuitive heuristic and it is used frequently \cite{Salathe2010}. It was found empirically to provide good results \cite{Sambaturu}. \textbf{(4) Eigenvalue centrality.} Another common approach it to select nodes using spectral graph properties, such as the eigenvalue centrality (e.g., \cite{Preciado2014,Yang2020}).

The main drawback of these heuristic algorithms is that they do not exploit all available information about dynamics. Specifically, they do not use negative test results, which contain information about the true distribution of the epidemic over network nodes.

\textbf{B. Supervised learning.} Algorithms that learn the risk per node using features of the temporal graph, its connectivity and infection state. Then, $k$ nodes with the highest risk are selected.
\textbf{(5) Supervised (vanilla).}
We treat each time step $t$ and each node $v_i$ as a sample, and train a 3-layer deep network using a cross entropy loss against the ground truth state of that node at time $t$. The input of the DNN has two components: A static component described in \secref{sec:ranking_module}, and a dynamic part that contains the number of infected neighbors and their neighbors (like \#1 above). Note that the static features include the, amongst other features, the degree and eigenvector centralities.  Therefore, if learning is successful, this baseline may derive an improved use of centralities based on local epidemic information.
\textbf{(6) Supervised (+GNN).}
Like \#5, but the input to the model is the set of all historic interactions of $v_i$'s and its $d$-order neighbours and their time stamps as an edge feature. The architecture is a GNN that operates on node and edge features. We used the same ranking module as our GNN framework, but the output probability is regarded as the probability that a node is infected.  
\textbf{(7) Supervised (+weighted degree).}
Same as \#6, but the loss is modified and nodes are weighted by their degree. Indeed, we wish to favour models that are more accurate on high-degree nodes, because they may infect a greater number of nodes.
\textbf{(8) Supervised (+weighted degree +GNN).}
Like \#6 above, using degree-weighted loss like \#7.

The supervised algorithm is trained to optimize the model $M$ by
minimizing the cross entropy loss function at every time step $t$ between the predicted probability that node $v$ will be infected $\hat{y}_v(t)$ to its groundtruth state ${y}_v(t)$ :
\begin{equation}
-\frac{1}{N}\sum_v w_v\ CrossEntropy({y}_v(t),\hat{y}_v(t))
\end{equation}

In the unweighted cross entropy loss, all nodes are weighted equally, $w_{v}=1$. We can strengthen this baseline by noting that we would like the model to be more accurate on high degree nodes, as they may infect a greater number of nodes. Hence, we weigh the contribution of a node to the loss term by its degree, $w_{v}=\deg(v)$. We refer to the latter algorithm as a degree weighted SL.

The main drawback of the supervised algorithms is that they optimize a myopic loss function. Therefore, they are unable to optimize the long term objective and consider the strategic importance of nodes.

\textbf{Metrics}. 
The end goal of quarantining and epidemiological testing is to minimize the spread of the epidemic. As it is unreasonable to eradicate the epidemic using social distancing alone, the hope is to ``flatten the curve'', namely, to slow down the epidemic progress. Equivalently, for a simulation with fixed length, the goal is to reduce the number of infected nodes. In addition to the \textbf{\%healthy} metric, defined in Sec. \ref{sec: Experiments}, we consider an additional metric, \textbf{\%contained}: The probability of containing the epidemic. This was computed as the fraction of simulations having cumulative infected nodes smaller than a fraction $\alpha$. We focus on this metric because it captures the important notion of the capacity of a health system. In the 2-community setup, where each community has half of the nodes, a natural choice of $\alpha$ is slightly greater than $0.5$, capturing those cases where the algorithm contains the epidemic within the infected community. In all the experiments we set $\alpha=0.6$. The only exception is the three-communities experiments, in which we set the bar slightly higher than $1/3$, and fixed $\alpha=0.4$.

\subsection{Epidemic Test Prioriritzation - Additional Experiments}

\begin{figure}

    \begin{tabular}{ccc}
    (a) GamesecRO & (b) Email & (c) ca-GrQc\\
    \includegraphics[width=0.31\columnwidth]{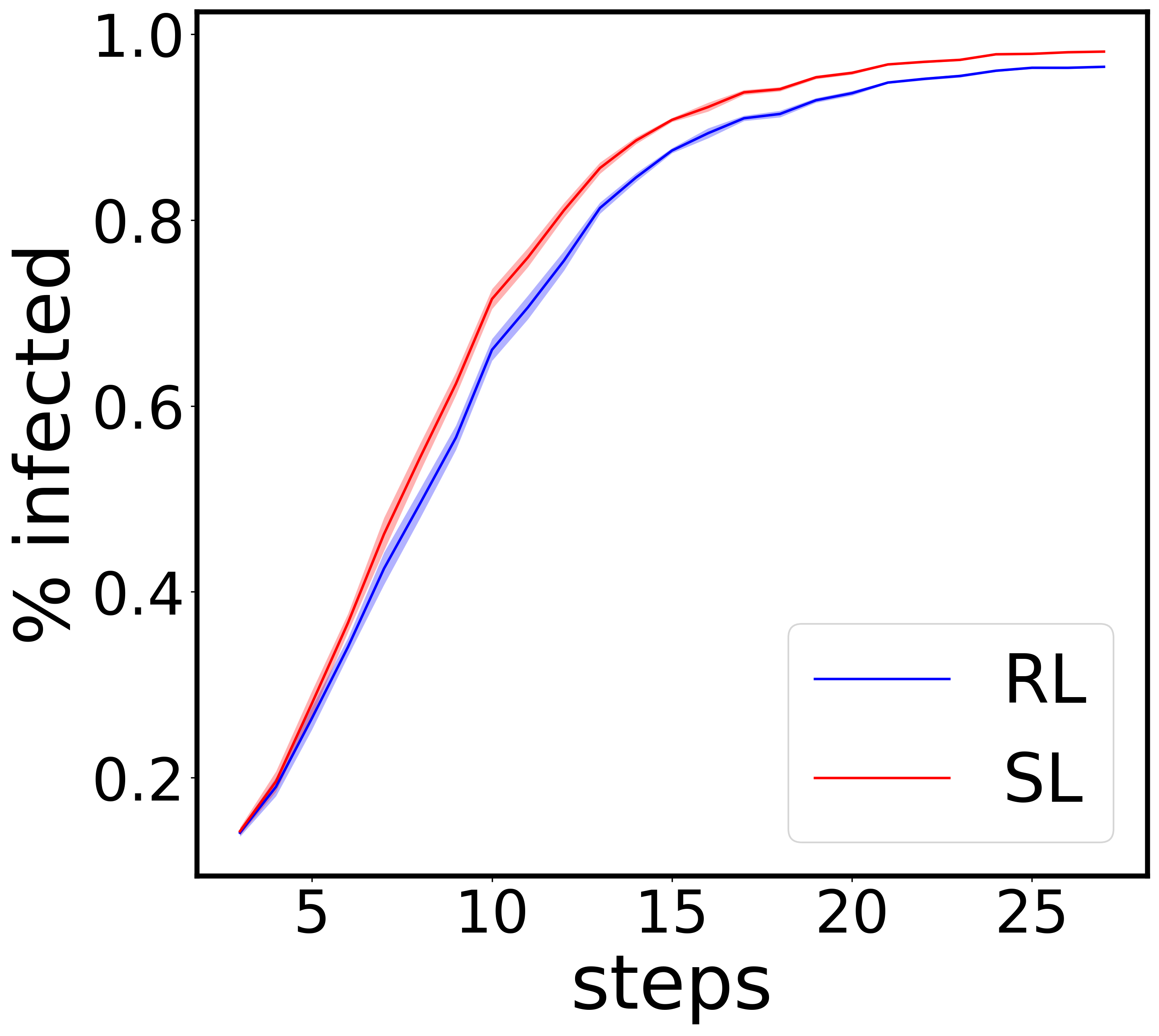}     &
    \includegraphics[width=0.31\columnwidth]{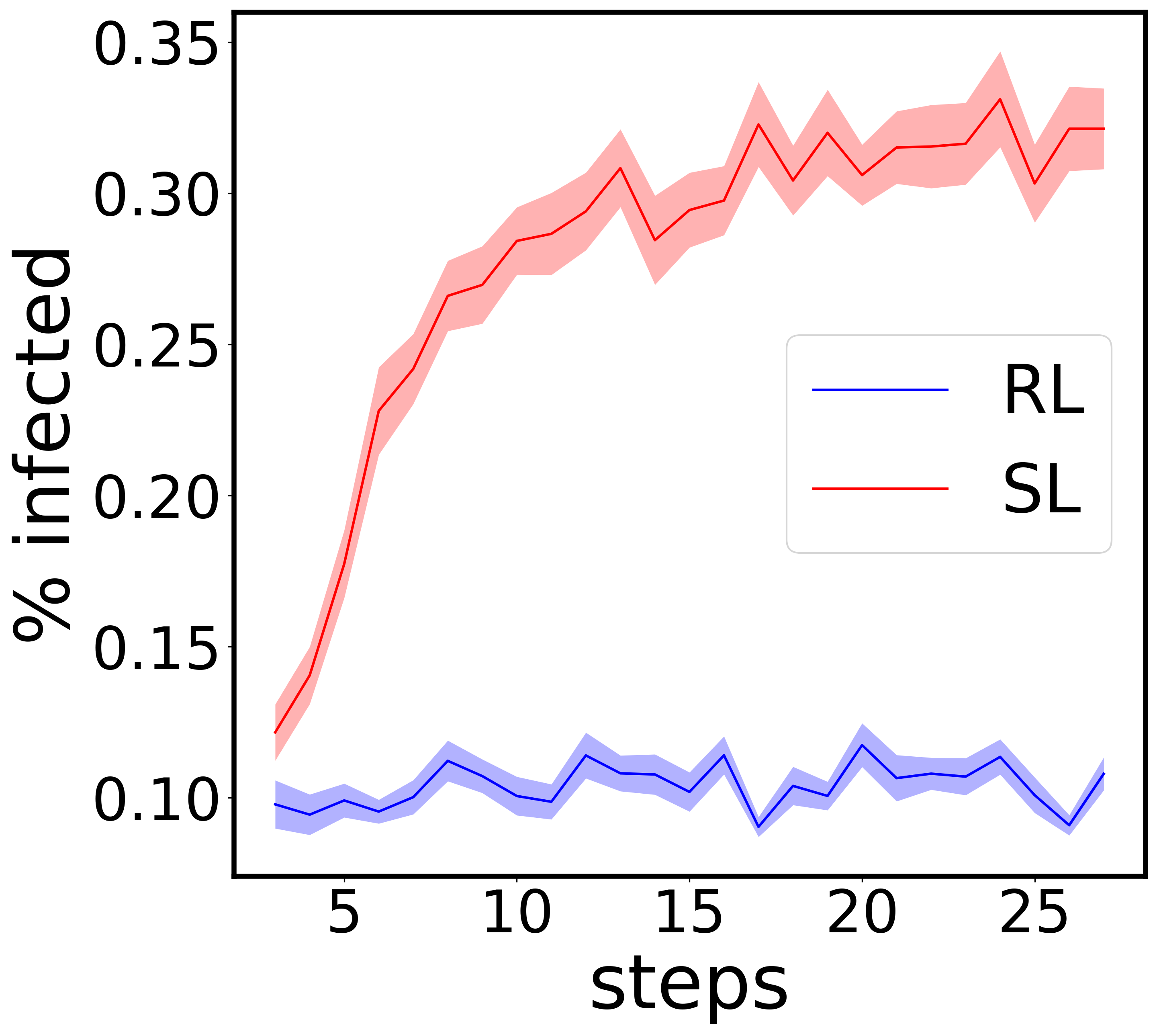}     &
    \includegraphics[width=0.31\columnwidth]{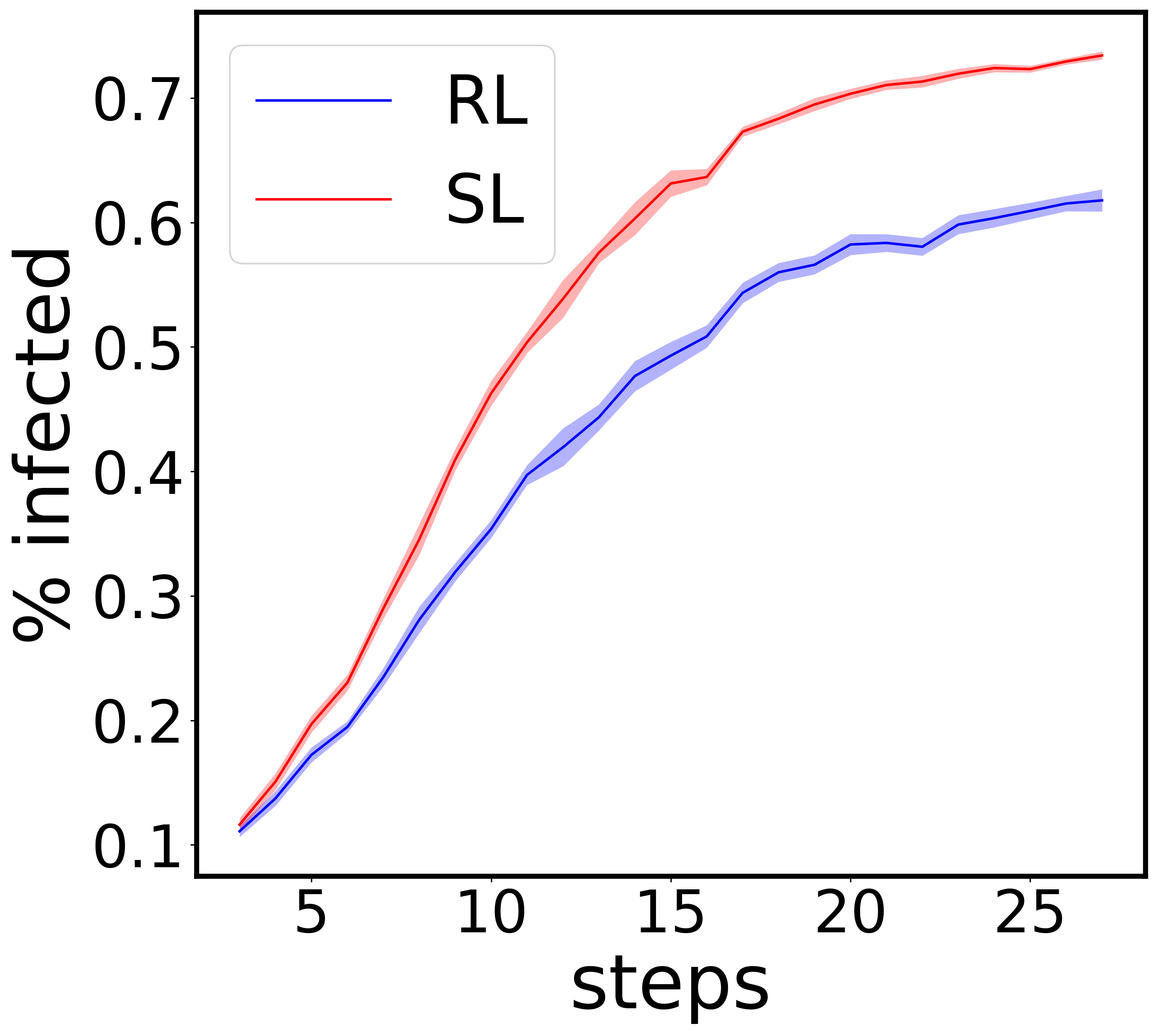}  
    \end{tabular}
    \caption{The fraction of infected nodes as a function of time step $t$. Shaded areas represent one standard deviation around the mean. (a) The epidemic propagation on an online social (gemsec-RO) (b)  The epidemic propagation on an email network. (c) The epidemic propagation on the collaboration graph ca-GrQc. In all cases the epidemic propagates more slowly under RLGN compared with the best baseline (supervised+GNN, \#4).
    \label{fig:epidemic-speed}
    }
\end{figure}

\textbf{Epidemic slowdown. }We investigated the progression of the epidemic under either RLGN or supervised+GNN algorithms. Figure \ref{fig:epidemic-speed} shows that the epidemic spread speed is slower under the RLGN policy in all graphs. In general, there are two extreme configuration regimes. First, the ``too-hard" case, when the number of tests is insufficient to block the epidemic, and second, the ``too-easy" case when there is a surplus of tests such that every reasonable algorithm can contain it. The more interesting case is the intermediate regime, where some algorithms succeed to delay the epidemic progression, or block it completely, better than other algorithms. Fig. \ref{fig:epidemic-speed}(a) illustrates the  case where the number of tests is insufficient for containing the epidemic, for all algorithms we tested. In Fig. \ref{fig:epidemic-speed}(b), the number of tests is insufficient for SL to block the epidemic. However, with same number of tests, RLGN algorithm  successfully contains the epidemic. Fig. \ref{fig:epidemic-speed}(c) presents an interesting case where RLGN slows down the epidemic progression and reduces the number of total number of infected node, compared with SL, but RL does not contain it completely.

\textbf{Epidemiological model variations.} \figref{fig:stability}(b) depicts a robustness analysis of \RLGN{} for variations in the epidemiological model. One of the most difficult quantities to assess is the probability for infection per social interaction. \figref{fig:stability}(b) shows that the trained model can sustain up to $\sim40\%$ deviation at test time in this key parameter.

\textbf{Graph size variations.} We have tested the robustness of our results to the underlying graph size. Specifically, we compare the two best algorithms RLGN (\#8) and SL+GNN (\#4), using graphs with various sizes, from 300 nodes to 1000 nodes. 
Table \ref{tab:300-nodes-pa} compares  RLGN with the SL+GNN algorithm on preferential attachment (PA) networks (mean degree $=2.8$). We provide results for various sizes of initial infection  $i_{0}$ and number of available tests $k$ at each step. The experiments show that there is a considerable gap between the performance of the RL and the second-best baseline.
Furthermore, RLGN achieves better performance than the SL+GNN algorithm with 40\%-100\% more tests. Namely, it increases the effective number of tests
by a factor of $\times1.4-\times2$.

\textbf{Initial infection size.} We also tested the sensitivity of the results to the relative size of the initial infection.
Table \ref{tab:300-nodes-pa} shows results when 4\% of the the network was initially infected, as well as for 7.5\% and 10\%. The results show that RLGN outperforms the baselines in this wide range of infection sizes.  

\subsection{Influence Maximization - Additional Experiments}

We presented two natural extensions to the Independent Cascades model. In Table \ref{tab:IM-geometric} in the main paper we presented the results of the IC(geometric) model. Table \ref{tab:IM_fixed-prob} presents the results of the IC(constant) model. In both cases, RLGN often outperform the state-of-the-art algorithms. In both simulations, the neighbor influence probability was $p=0.25$, the agent's success probability for setting a node to \emph{Influenced} state was $q=0.3$, and the probability that an influenced node will reveal its state was $\eta=0.25$.

We follow with a comparison of the performance of the various algorithms on the Linear Threshold model (Table \ref{tab:IM_LT}). Here, RLGN clearly outperformed the other baselines. The main reason to the increased gap is that a Linear Threshold  model contains more parameters, as each edge and  is associated with a random weight and each node is assigned a peer resistance value. RLGN, as a trainable model, is able to uncover relevant patterns, while the other algorithms fail to do so. Table \ref{tab:IM_LT} also shows that the RLGN performs better thatn the baselines over a variety of the dynamic model parameters.

\begin{table}
\centering
\begin{tabular}{|l|c|c|c|c|}
\toprule
{} & gemsec-RO & Email & ca-HepTh &  ca-GrQc \\
\midrule
LIR          &       $37.5\pm 1$ &     $73.2 \pm 0.3$ &      $58.0 \pm 0.3$  &  $47.4 \pm 0.4$\\
\hline 
LIR (filtered) &       $36.3\pm 1$ &     $73.6 \pm 0.3$ &      $58.3 \pm 0.4$ &  $48.7 \pm 0.4$\\
\hline 
Degree       &       $25.4\pm 1$ &     $74.0 \pm 0.3$ &       $58.0 \pm 0.3$ &  $48.5 \pm 0.4$\\
\hline 
Degree Discounted  &       $25.9\pm 1$ &     $70.4 \pm 0.5$ &      $58.2 \pm 0.5$ &  $48.4 \pm 0.4$\\
\hline 
Eigenvector  &       $24.8\pm 1$ &     $7\mathbf{4.3 \pm 0.3}$ &       $48.6 \pm 0.6$ &  $48.5 \pm 0.4$ \\
\hline 
RLGN         &       $\mathbf{71.6\pm 1}$ &     $\mathbf{74.2 \pm 0.2}$ &      $\mathbf{59.5 \pm 0.5}$ &  $48.9 \pm 0.3$ \\
\bottomrule
\end{tabular}
\caption{\label{tab:IM_fixed-prob} The percentile of influenced nodes on the real-world networks in the Influence Maximization setup. IC(constant) was used as the dynamical model. Each episode lasted $15$ steps, and each result represents the mean \%\emph{influenced} value after $300$ episodes. Results on the Portland network were omitted as all algorithms were able to achieve $>98\%$ influenced share on this network.}
\end{table}

\begin{table}
\centering
\begin{tabular}{|l|c|c|c|c|}
\toprule
{} &  ca-GrQc & gemsec-RO &  ca-HepTh & Email \\
\midrule
LIR          &  $3.8\pm0.1$ &     $0.014\pm0.006$ &   $1.4\pm0.1$ &      $14.2\pm0.4$ \\
\hline 
LIR(filtered) &  $5.2\pm0.1$ &     $0.019\pm0.001$ &   $3.1\pm0.2$ &      $15.8\pm0.3$ \\
\hline 
Degree       &  $4.5\pm0.1$ &    $0.007\pm0.002$ &   $1.4\pm0.1$ &     $17.3\pm0.3$ \\
\hline 
Degree Discounted           &  $4.7\pm0.1$ &    $0.007\pm0.004$ &  $1.14\pm0.1$ &     $14.0\pm0.4$ \\
\hline 
Eigenvector  &  $4.7\pm0.1$ &    $0.008\pm0.004$ &  $0.42\pm0.01$ &     $18.4\pm0.3$ \\
\hline 
RLGN                     &  $\mathbf{7.1\pm0.2}$ &       $\mathbf{1.98\pm0.06}$ &   $\mathbf{5.7\pm0.2}$ &     $\mathbf{21.8\pm0.1}$ \\
\bottomrule
\end{tabular}

\vspace*{0.2 cm} 

\begin{tabular}{|l|c|c|c|c|}
\toprule
{} &  ca-GrQc  & gemsec-RO &  ca-HepTh & Email \\
\midrule
LIR          &    $5.5\pm 0.1$&      $0.028\pm 0.001$ &   $2.8\pm 0.2$ &   $18.9\pm 0.5$  \\
\hline 
LIR(filtered) &    $6.9\pm 0.1$ &     $0.034\pm 0.002$ &   $5.6\pm 0.2$ &   $21.2\pm 0.4$  \\
\hline 
Degree      &    $6.2\pm 0.1$  &     $0.01\pm 0.001$ &   $2.8\pm 0.2$ &  $23.6\pm 0.3$ \\
\hline 
Degree Discounted           &     $6.1\pm 0.1$  &    $0.009\pm 0.001$ &   $3.2\pm 0.2$ & $18.9\pm 0.4$  \\
\hline 
Eigenvector  &     $6.2\pm 0.1$ &     $0.011\pm 0.001$ &  $0.58\pm 0.03$ & $24.2\pm 0.3$  \\
\hline 
RL+GNN                     &      $\mathbf{10.7\pm 0.1}$ &       $\mathbf{4.3\pm 0.1}$ &   $\mathbf{12.2\pm 0.2}$ &  $\mathbf{26.5\pm 0.1}$  \\
\bottomrule
\end{tabular}

\caption{\label{tab:IM_LT} The percentile of influenced nodes on the real-world networks in the Influence Maximization setup. Linear threshold was used as the dynamical model. In the top table, the peer resistance value $z_v$ was sampled uniformly from $[0.5,1.5]$ and in the lower table it was sampled uniformly from $[0.4,1.4]$. Each edge was assigned a uniform random weight $[0,1]$. Each episode lasted $20$ steps, and each result represents the mean \%\emph{influenced} value after $300$ episodes.}
\end{table}

\subsection{Ablation Studies}

\paragraph{Mapping scores to action distribution.} 

\begin{figure}
    \centering
    \includegraphics[width=0.48\columnwidth]{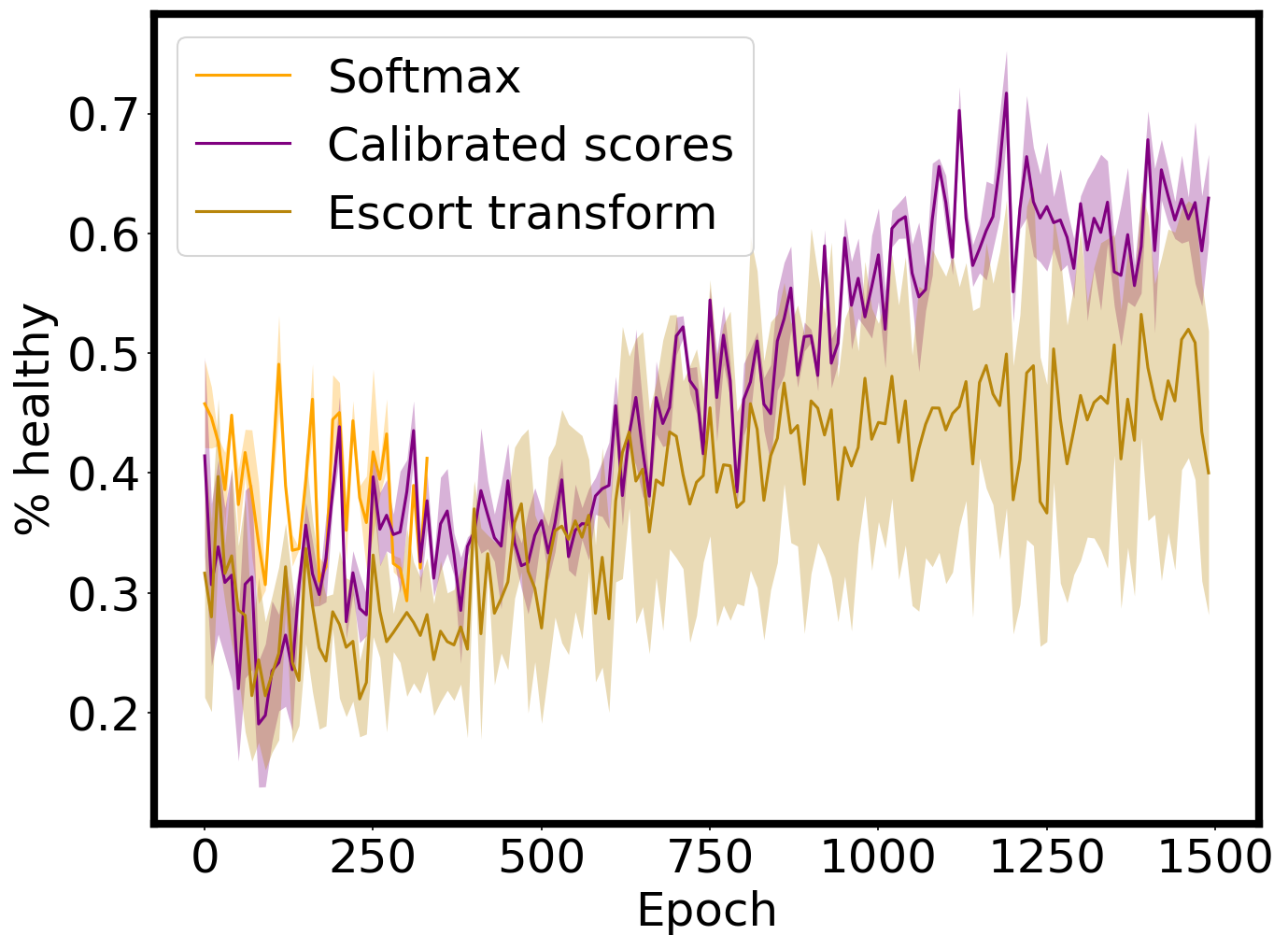}
    \includegraphics[width=0.48\columnwidth]{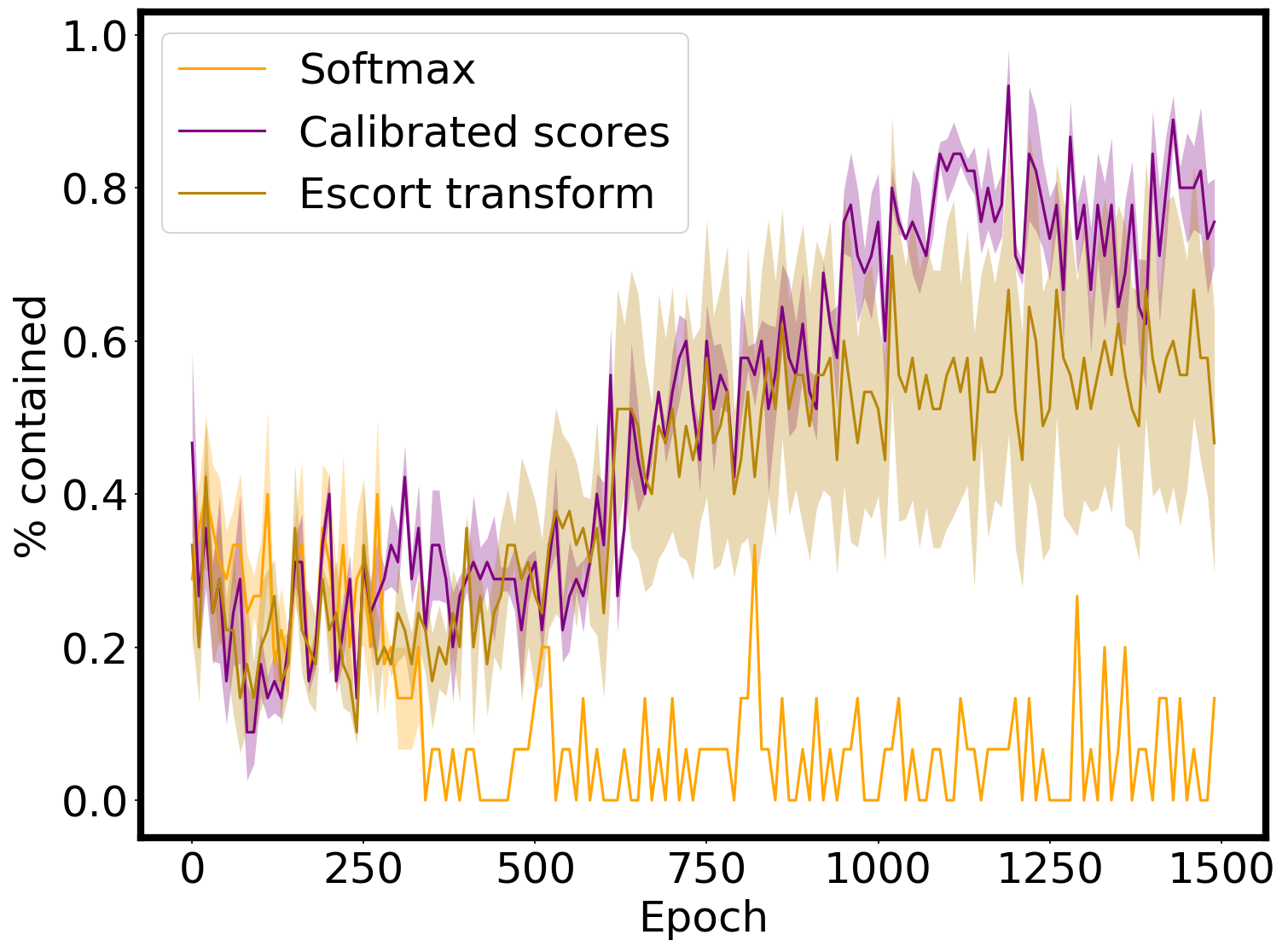}
    \caption{ Fraction of contained epidemics and \%healthy during training in a preferential attachment model with $200$ nodes and a mean degree $2.8$. For non-normalized mapping, only one of the three seeds in the softmax distribution simulation completed training due to numerical instability. No stability issues were observed when using  the calibrated scores normalization scheme described by \eqref{eq: score-to-prob}.
    \label{fig:score-to-prob}}
\end{figure}

We compare the performance of our score-to-probability function (calibrated-scores) to the popular softmax (Boltzmann) distribution. In practice, in most instances, we were unable to train a model using the softmax distribution as the neural network weights diverge. Fig.~\ref{fig:score-to-prob} presents the training curve in one of the few instances that did converge. It is clear that the model was not able to learn a useful policy while using the calibrated-scores probability function resulted in a corresponding value of more than $0.75$. We also compare our approach with the recent escort transform \cite{mei2020}, a state-of-the-art score-to-probability method. As shown in Fig.~\ref{fig:score-to-prob}, our method outperforms the escort transform in this problem.

\begin{table}
\centering
    \begin{tabular}{|c|c|c|}
        \hline 
         & $\%contained$ & \# training epochs\\
        \hline 
        \hline 
        Sigmoid & $0.84\pm0.05$ & 1210\\
        \hline 
        GRU & $0.91\pm0.03$ & 810\\
        \hline 
        $L_{2}$ norm. & $\mathbf{0.93\pm0.02}$ & \textbf{500}\\
        \hline 
    \end{tabular}
    \caption{\label{tab:norm-table} Training time and fraction of contained epidemic for three normalization schemes. The $L_2$ normalization scheme is fastest and achieves the best performance.}
\end{table}

\paragraph{Normalization in scale-free networks.} 

We compared the suggested normalization to a number of other alternative normalization methods. (1) Applying a sigmoid layer after the hidden state update module $G$. (2) Replace the hidden state update module with a GRU layer. (3) Apply $L_2$ normalization to each feature vector $h_v(t)$ (similarly to \cite{Hamilton2017}) (4) Normalize the feature vector matrix by its $L_2$ norm.
These four normalization schemes span three different types of normalization: single-feature normalization (1+2), vector normalization (3), and matrix normalization (4).

Table \ref{tab:norm-table} presents the score after training and the number of training steps required to complete training. Method (4) was unstable and training did not converge, therefore it was omitted from the table.  The main reason for the training time difference is that without normalization, the DNN weights' magnitude increases. In a GRU module, or with a direct application of a sigmoid layer, the features pass through a sigmoid activation function. When the  magnitude of the input to this layer is large, the gradient is very small due to the sigmoid plateau. This substantially slows down the learning process. 


\begin{figure}
\centering
\includegraphics[width=0.6\columnwidth]
{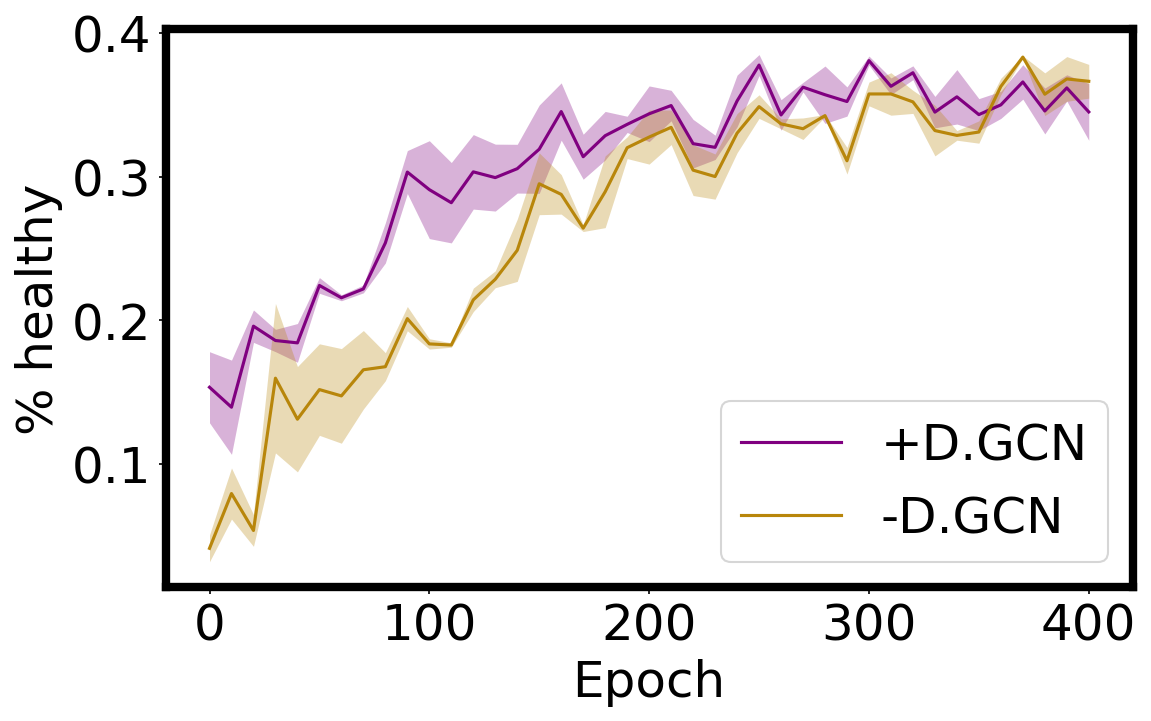}
\caption{Training curves with and without a local-diffusion GCN (D. GCN) module on a preferential attachment network. \label{fig:local_diffusion}}
\end{figure}

\begin{table}
\centering
\begin{tabular}{|l|c|c|}
\hline 
 & \#Nodes & \#Edges\\
\hline 
\hline 
CA-GrQc & 5242 & 14496\\
\hline 
Montreal & 103425 & 630893\\
\hline 
Portland & 10000 & 199167\\
\hline 
Email & 32430 & 54397\\
\hline 
GEMSEC-RO & 41773 & 222887\\
\hline 
\end{tabular}\caption{\label{tab:datasets-info}Number of edges and nodes in the large-scale
datasets}
\end{table}

\begin{table}
\centering
    \setlength{\tabcolsep}{3pt}
    \scalebox{0.97}{{\sc{
        \begin{tabular}{|l|c|c|}
        \hline 
         \%contained & 2 communities & 3 communities \\ \hline 
        \hline 
        Tree-based model & $15\pm 35$  &  $0\pm0$  \\ \hline 
        Counter model & $19\pm39$  & $ 1\pm 4$ \\ \hline 
               
        Degree centrality  & $23\pm1$  &  $0\pm0$   \\ \hline
        Eigenvector centrality & $19\pm3$ & $0\pm0$\\  \hline 
        Supervised (vanilla) & $24\pm 11$ & $2\pm 2$ \\ \hline 
        Supervised +GNN & $27\pm 10$  & $ 2\pm 2$  \\ \hline 
        Supervised +degree & $29\pm 10$  & $ 1\pm 2$\\ \hline 
        Supervised +GNN+deg & $24\pm 10$ & $2\pm 02$\\ \hline 
        \RLGN{} (Vanilla) & $66\pm 10$  & $7\pm 5$   \\ \hline 
        \RLGN{} full (ours) & \textbf{88}{$\boldsymbol{\pm}$}\textbf{1}  & \textbf{53}{$\boldsymbol{\pm}$}\textbf{13} \\
        \hline 
        \end{tabular}
    }}}
    \caption{Probability (in \%) of containing an epidemic in community-based networks. Each community has $30$ densely connected nodes, and the test budget is $k=2$.
    }
    \label{tab:comaprison-chart}
\end{table}


\begin{table}
\centering
    \setlength{\tabcolsep}{3pt}
    \begin{tabular}{|l|c|c|c|c|c|c|}
    \hline 
    \multirow{2}{*}{\textbf{$n=300$}} & \multicolumn{2}{c|}{Init. infection size 5\%} & \multicolumn{2}{c|}{Init. infection size 7.5\%} & \multicolumn{2}{c|}{Init. infection size 10\%}\\
    \cline{2-7} \cline{3-7} \cline{4-7} \cline{5-7} \cline{6-7} \cline{7-7} 
     & \%healthy & \%contained & \%healthy & \%contained & \%healthy & \%contained\\
    \hline 
    SL, $k=1\%$ & $27\pm2$ & $15\pm5$ & $21\pm2$ & $4\pm2$ & $18\pm1$ & $1\pm1$\\
    \hline 
    SL, $k=1.33\%$ & $41\pm3$ & $37\pm6$ & $27\pm2$ & $12\pm4$ & $24\pm2$ & $6\pm3$\\
    \hline 
    SL, $k=2\%$ & ${66\pm4}$ & $76\pm6$ & ${48\pm3}$ & $55\pm7$ & $37\pm2$ & $32\pm6$\\
    \hline 
    RLGN, $k=1\%$ & $50\pm2$ & ${78\pm7}$ & {$43\pm2$} & ${58\pm1}$ & ${40\pm1}$ & ${48\pm6}$\\
    \hline 
    \end{tabular}
    
    \vspace*{0.2 cm} 

    \begin{tabular}{|l|c|c|c|c|c|c|}
    \hline 
    \multirow{2}{*}{\textbf{$n=500$}} & \multicolumn{2}{c|}{Init. infection size 5\%} & \multicolumn{2}{c|}{Init. infection size 7.5\%} & \multicolumn{2}{c|}{Init. infection size 10\%}\\
    \cline{2-7} \cline{3-7} \cline{4-7} \cline{5-7} \cline{6-7} \cline{7-7} 
     & \%healthy & \%contained & \%healthy & \%contained & \%healthy & \%contained\\
    \hline 
    SL, $k=1\%$ & $24\pm2$ & $7\pm4$ & $20\pm1$ & $2\pm1$ & $19\pm1$ & $0\pm1$\\
    \hline 
    SL, $k=1.6\%$ & $48\pm3$ & $54\pm6$ & $35\pm2$ & $27\pm7$ & $29\pm1$ & $11\pm1$\\
    \hline 
    SL, $k=2\%$ & ${67\pm3}$ & $83\pm5$ & ${46\pm2}$ & $53\pm4$ & $38\pm2$ & $37\pm7$\\
    \hline 
    {RLGN, ${k=1\%}$} & $52\pm1$ & ${97\pm2}$ & ${44\pm2}$ & ${75\pm11}$ & ${42\pm1}$ & ${66\pm6}$\\
    \hline 
    \end{tabular}

    \vspace*{0.2 cm} 

    \begin{tabular}{|l|c|c|c|c|c|c|}
    \hline 
    \multirow{2}{*}{\textbf{$n=1000$}} & \multicolumn{2}{c|}{Init. infection size 5\%} & \multicolumn{2}{c|}{Init. Infection size 7.5\%} & \multicolumn{2}{c|}{Init. infection size 10\%}\\
    \cline{2-7} \cline{3-7} \cline{4-7} \cline{5-7} \cline{6-7} \cline{7-7} 
     & \%healthy & \%contained & \%healthy & \%contained & \%healthy & \%contained\\
    \hline 
    SL, $k=1\%$ & $25\pm2$ & $5\pm3$ & $21\pm1$ & $0\pm1$ & $19\pm1$ & $0\pm0$\\
    \hline 
    SL, $k=1.5\%$ & $42\pm2$ & $49\pm6$ & $30\pm1$ & $10\pm3$ & $27\pm1$ & $4\pm2$\\
    \hline 
    SL, $k=2\%$ & ${66\pm1}$ & $84\pm5$ & ${45\pm2}$ & $59\pm5$ & $37\pm1$ & $30\pm1$\\
    \hline 
    {RLGN, ${k=1\%}$} & $52\pm1$ & ${97\pm2}$ & ${44\pm2}$ & ${75\pm11}$ & ${42\pm1}$ & ${66\pm6}$\\
    \hline 
    \end{tabular}
    \caption{A comparison between RLGN and SL+GNN (the best epidemic test prioritization baseline). RLGN performance is highlighted. The number of additional resources needed to surpass the RLGN performance in a given metric is also highlighted.  In many cases, even using SL+GNN with twice as many resources than RLGN performs worse than RLGN. The evaluation was performed on a preferential attachment network with mean degree $2.8$. The number of nodes is indicated at the top of each table.}
    \label{tab:300-nodes-pa}
\end{table}

\subsection{Network architecture}
The architecture of the ranking module  is shared by algorithms \#4, \#6 and \#8 with slight variations indicated below.

\paragraph{Input.} We encode the dynamic node features $\zeta_{v}^{d}(t)$
as follows. \textbf{Epidemic test priorization}: A one hot hot vector of dimension $4$. Each of the first three elements corresponds to one of the three mutually exclusive options, which depends on the previous step: untested, tested positive, tested negative. The last entry indicates whether a node
was found positive in the past. \textbf{Influence maximization}: A one hot hot vector of dimension $3$. Similarly, the first two elements indicate whether a node has indicates it is influenced in the previous step or not, and the last element whether it indicated so in the past.
The static node features, $\zeta_{v}^{s}(t)$, are common to both problems.
As described in the main paper, topological graph centralities (betweenness,
closeness, eigenvector, and degree centralities) and random node features. The graph centralities are standard metrics, and were calculated using NetworKit. An ablation study showed a maximal performance difference of 2\% between different centralities subsets and was statistically insignificant.

\paragraph{Local Diffusion GNN. }This module $M_{e}$ is composed of a single
graph convolutional layer. The input features are  the last time
step node features. The number of output features is 64. 

\paragraph{Long Range Information GNN. }Each message passing module $M^{l}$ contains one hidden layer, where the number of hidden features is 64. After both the hidden and the last layer we apply a leaky ReLu layer with leakage constant $0.01$. After aggregating the result using  the addition aggregation function, we apply an additional MLP with one layer (linear+ReLu) on the resulting feature vector. The number of output features is 64. The value of $\tau$, the information window size, was 7 in all our experiments.

We experimented with the numbers of stacked modules $l$ (layers). We found that $l=3$ performed slightly better than $l=2$ but training was considerably slower because the batch size had to be reduced. We therefore used $l=2$ in all experiments reported.

\paragraph{Hidden state update.} The hidden state MLP $G$ is composed of a single linear layer follows by a ReLu activation layer. To keep the resulting hidden feature vector (of dimension 64) norm under check, an additional
normalization scheme is then applied. This module
was replaced by a GRU layer in the ablation studies.

\paragraph{Output layer.} The last module is a single linear layer, with an output dimension as the number of the nodes in the graph.

\paragraph{Learning framework.} We used Pytorch \cite{paszke2017automatic} and Pytorch Geometric \cite{fey2019fast} to construct the ranking module. We used ADAM with default parameters as our optimizer.

\subsection{Training protocol}
\textbf{Initializaition}. The initialization depends on the problem setup.  Each influenced node signal with probability $q$ at each turn.
In the epidemic tests prioritization setup we trained the RL and SL algorithms by generating random networks and initializing each network by selecting for each instance a random subset of $m_{0}$  infected nodes. We propagate the epidemic until it spans at least $i_0$ infected nodes (for at least $t_0$ steps), and randomly detect a subset of the infected nodes of size $k_0<i_0$. The initialization for the Influence Maximization setup is simpler, and conclude in generating a random network. 

At each step, in all algorithms but RL, we pick the top $k$ rated nodes. In RL, we perform the same procedure during the evaluation phase, while during training we sample $k$ nodes using the score-to-probability distribution. 

Each model was trained for at most 1500 episodes, but usually, training was completed after 1000 episodes. Each episode contained 1024 steps,
collected by 4 different workers. As our network contains a recurrent module, we propagate each sample in the buffer for three steps, in
a similar fashion to R2D2.

For each setup we described, at least three models were trained using different seeds, and the results are the average over the performance of all models. The errors are the standard deviation of the mean. over at least 100 evaluation episodes for each model.

Each episode lasted for 25 steps, each corresponds conceptually to a day. The transition time from the latent to the infectious state was normally distributed with a mean of two steps and a standard deviation
of 1 step, corresponding to real-world values. The advantage was calculated using the Generalized Advantage framework with parameters $\gamma=0.99,\lambda=0.97.$

Table \ref{table:parameters}  presents the simulation parameters used in the main paper.
We shall make the repository and code available online.

\begin{table}
\centering
    \begin{tabular}{|l|c|}
    \hline 
    minimal \#propagation steps ($t_{0}$) & 4\\
    \hline 
    \multirow{3}{*}{minimal \#infected component size ($i_{0}$)} & communities: 4 (same community)\\
    \cline{2-2} 
     & preferential attachment: 5\%\\
    \cline{2-2} 
     & contact tracing: 7\%\\
    \hline 
    Learning rate & $3\cdot10^{-4}$\\
    \hline 
    $\lambda$ & 0.97\\
    \hline 
    $\gamma$ & 0.99\\
    \hline 
    Entropy loss weight & 0.01\\
    \hline 
    Value loss weight & 0.5\\
    \hline 
    Probability distribution of $e\in\mathcal{E}(t)$ & $U[0.2,0.6]$\\
    \hline 
    Batch size & 256 (128 if \#nodes$>200$)\\
    \hline 
    3-communites SBM matrix & $\left(\begin{array}{ccc}
    0.6 & 0.001 & 0\\
    0.001 & 0.6 & 0.001\\
    0 & 0.001 & 0.6
    \end{array}\right)$\\
    \hline 
    2-communites SBM matrix & $\left(\begin{array}{cc}
    0.6 & 0.0022\\
    0.0022 & 0.6
    \end{array}\right)$\\
    \hline 
    \end{tabular}
    \caption{Parameters table}
    \label{table:parameters}
\end{table}

\section{The tree model baseline \label{sec:tree model}}
In this appendix We describe our tree model baseline (algorithm \#1). Consider an epidemic propagating on a tree, and assume there is a single initially infected node ("patient zero"). In this case, there is a single path from the infection source to every node in the graph and we can we can analytically solve for the probability a node is infected, given that the root of the tree was infected at time $t_0$. This model is useful when the underlying network is locally a tree, i.e, that for every new infected node $v$ there is w.h.p just one node which may have infected it.

We start with a simple case.

\subsection{Simple case: No latent state}

Let us first consider a simple model in which the epidemic spreads
on a tree like structure with a single epidemic source, a.k.a. patient-zero,
as the root. For now, let us assume there is no latent state. 

Our goal is to calculate the probability that a node $n$ will
be infected at time $T$

\[
F_{n}(T)  \triangleq \Pr\left(ST_{n}(T)=\mathcal{\mathcal{I}}|ST_{r}(0)=\mathcal{I}\right)
\]

For every node $j$ there is a single path from the node to the root,
denoted by $r$. Let us assume the path is
$\{y_{0}=r,y_{1},y_{2},..y_{n-1},y_{n}=j\}$. Assume that in $[0,T]$  a sequence
of interactions between node $y_{n}$ and $y_{n-1}$ occurred
at discrete times $(t_{1},t_{2},...t_{m})$, and that each interaction
is characterized by an infection probability $(p_{1},p_{2},...p_{m})$.
We  evaluate $F_{n}(T)$ by induction. For abbreviation, we write $ST_{y_{i}}(t)=Y_{i}(t)$
and denote the event $ST_{r}(0)=\mathcal{I}$ as $A$. 

Our key result is that The state of node
$n$ at the time of interaction $m$ is a function of its state at
penultimate interaction time $F_{n}(t_{m-1})$, the interaction transmission
probability $p_{m}$, and the predecessor node $n-1$ state at time
$m$, $F_{n}(t_{m-1})$.

\begin{align*}
    F_{n}(t_{m}) & =F_{n}(t_{m-1})+p_{m}\left(F_{n-1}(t_{m})-F_{n}(t_{m-1})\right)\\
     & =p_{m}F_{n-1}(t_{m})+F_{n}(t_{m-1})\left(1-p_{m}\right)
\end{align*}

The first term is the probability to get infected at the $m$ interaction,
and the second term is the probability to get infected before hand. We shall now prove this result.

\begin{proof}
 We can write the conditional probability using a graphical model decomposition and obtain 
\begin{align}
    & \Pr\left(Y_{n}(T)=\mathcal{\mathcal{I}}|A\right) = & \nonumber \\
    & \Pr\left(Y_{n}(t_{m})=\mathcal{\mathcal{I}}|Y_{n-1}(t_{m})=\mathcal{\mathcal{I}},A\right)\Pr\left(Y_{n-1}(t_{m})=\mathcal{\mathcal{I}}|A\right) =& \label{eq:conditional on previous}\\
    & \Pr\left(Y_{n}(t_{m})=\mathcal{\mathcal{I}}|Y_{n-1}(t_{m})=\mathcal{\mathcal{I}},A\right)F_{n-1}(t_{m})\nonumber 
\end{align}
since if the ancestor node is not in an infectious state, the decedent can not be infected. Denote the indicator that interaction $l$ was able to transmit the epidemic as $I_{l}$. We have,
\begin{align*}
    & \Pr\left(Y_{n}(t_{m})=\mathcal{\mathcal{I}}|Y_{n-1}(t_{m})=\mathcal{\mathcal{I}},A\right) & =\\
    & \sum_{l=1}^{m}\Pr\left(\text{\ensuremath{y_{n}}'s infection time is \ensuremath{t_{l}}}|Y_{n-1}(t_{m})=\mathcal{\mathcal{I}},A\right) & =\\
    & \sum_{l=1}^{m}\Pr\left(Y_{n}(t_{l-1})=\mathcal{H},I_{l},Y_{n-1}(t_{l})=\mathcal{\mathcal{I}}|Y_{n-1}(t_{m})=\mathcal{\mathcal{I}},A\right)
\end{align*}

As, for an infection event to take place at it must be that node $y_{n-1}$ was infected at $t_{l}$, node $y_{n}$ was healthy beforehand, and that the interaction resulted in an infection. We can now write this
as 
\begin{align}
    & \Pr\left(Y_{n}(t_{l-1})=\mathcal{H},I_{l},Y_{n-1}(t_{l})=\mathcal{\mathcal{I}}|Y_{n-1}(t_{m})=\mathcal{\mathcal{I}},A\right) & =\nonumber \\
    & p_{l}\Pr\left(Y_{n}(t_{l-1})=\mathcal{H},Y_{n-1}(t_{l})=\mathcal{\mathcal{I}}|Y_{n-1}(t_{m})=\mathcal{\mathcal{I}},A\right) & =\nonumber \\
    & p_{l}\Pr\left(Y_{n}(t_{l-1})=\mathcal{H},Y_{n-1}(t_{m})=\mathcal{\mathcal{I}}|Y_{n-1}(t_{l})=\mathcal{\mathcal{I}},A\right)\frac{\Pr\left(Y_{n-1}(t_{l})=\mathcal{\mathcal{I}}|A\right)}{\Pr\left(Y_{n-1}(t_{m})=\mathcal{\mathcal{I}}|A\right)} & =\label{eq:time-shift}\\
    & p_{l}\Pr\left(Y_{n}(t_{l-1})=\mathcal{H}|Y_{n-1}(t_{l})=\mathcal{I},A\right)\frac{F_{n-1}(t_{l})}{F_{n-1}(t_{m})} & =\nonumber \\
    & p_{l}\left(1-\Pr\left(Y_{n}(t_{l-1})=\mathcal{I}|Y_{n-1}(t_{l})=\mathcal{I},A\right)\right)\frac{F_{n-1}(t_{l})}{F_{n-1}(t_{m})}\nonumber 
\end{align}

The transition from the first line to the second is due to the independence of the interaction infection probability with the history of the participating parties. The third line is Bayes' theorem. If a node is infected at time $t_{l}$, it will be infected later on at $t_{m}$, as expressed in line 4. The last line is the complete probability formula. 

We rewrite $\Pr\left(Y_{n}(t_{l-1})=\mathcal{I}|Y_{n-1}(t_{l})=\mathcal{I},A\right)$ as
\begin{align*}
    & \Pr\left(Y_{n}(t_{l-1})=\mathcal{I}|Y_{n-1}(t_{l})=\mathcal{I},A\right) =& \\
    & \frac{\Pr\left(Y_{n}(t_{l-1})=\mathcal{I}|A\right)-\Pr\left(Y_{n}(t_{l-1})=\mathcal{I},Y_{n-1}(t_{l})=\mathcal{H}|A\right)}{\Pr\left(Y_{n-1}(t_{l})=\mathcal{I}|A\right)} = & \\
    & \frac{\Pr\left(Y_{n}(t_{l-1})=\mathcal{I}|A\right)}{\Pr\left(Y_{n-1}(t_{l})=\mathcal{I}|A\right)} =& \\
    & \frac{F_{n}(t_{l-1})}{F_{n-1}(t_{l})}
\end{align*}
The transition from the first line to the second line is a complete probability transition. The third line is due to the fact that if $y_{n-1}$ was not infected at time $t_{l}$, clearly $y_{n}$ could not be infected before $t_{l}$. We have
\begin{align*}
    F_{n}(t_{m})=\Pr\left(Y_{n-1}(t_{m})=\mathcal{\mathcal{I}}|A\right) & =\sum_{l=1}^{m}p_{l}\left(1-\frac{F_{n}(t_{l-1})}{F_{n-1}(t_{l})}\right)\frac{F_{n-1}(t_{l})}{F_{n-1}(t_{m})}F_{n-1}(t_{m})\\
     & =\sum_{l=1}^{m}p_{l}\left(F_{n-1}(t_{l})-F_{n}(t_{l-1})\right)
\end{align*}

Therefore, given $F_{n-1}(t_{l})$ for all $l\in\{1..n-1\}$ and $F_{n}(t_{l})$
for all $l\in\{1..n\}$, we can directly calculate the infection probabilities, given the initial condition: $F_{i}(0)=\delta_{i,0}$.

We can write the partial density function of $F_{i}(t_{l})$ as $f_{i}(t_{l}) = F_{i}(t_{l})-F_{i}(t_{l-1})$, and obtain: $
f_{n}(t_{m}) = p_{m}\left(F_{n-1}(t_{m})-F_{n}(t_{m-1})\right)$. This allows us to write this with an intuitive formulation
\begin{align*}
    F_{n}(t_{m}) & =F_{n}(t_{m-1})+p_{m}\left(F_{n-1}(t_{m})-F_{n}(t_{m-1})\right)\\
     & =p_{m}F_{n-1}(t_{m})+F_{n}(t_{m-1})\left(1-p_{m}\right)
\end{align*}
The first term is the probability to get infected at the $m$ interaction, and the second term is the probability to get infected before hand.

\end{proof}

\subsection{Full analysis with latent states}
We now discuss the case where a node can be in a latent state. The main difference is that the complement of the infectious state
is composed of two states, healthy $\mathcal{H}$, and latent $\mathcal{L}$.
We shall denote all the non-infecting states as $\mathcal{H}^{+}=\{\mathcal{\mathcal{H}},\mathcal{L}\}$
and all the infected states as $\mathcal{I}^{+}=\{\mathcal{\mathcal{I}},\mathcal{L}\}$,
and sometime abuse the notation by writing $S_{i}(t)=\mathcal{H}^{+}$.
We denote the transmission delay from the latent to infectious state as $L(\tau)$.

As before, we are interested in the probability that 
\[
\Pr\left(Y_{n}(T)=\mathcal{\mathcal{I}^{+}}|S_{r}(0)=\mathcal{I}\right)
\]
The derivation below shows that, similar to the previous case, we
can solve for this probability using dynamic programming. The end result is that  
\begin{align*}
    \Pr\left(Y_{n}(T)=\mathcal{\mathcal{I}^{+}}|ST_{r}(0)=\mathcal{I}\right) & =\sum_{l=1}^{m}p_{l}\left(F_{n-1}(t_{l})-F_{n}(t_{l-1})-\Pr\left(Y_{n}(t_{l-1})=\mathcal{L}|A\right)\right)\, ,
\end{align*}
with 
\begin{align*} 
    \Pr\left(Y_{n}(t_{l}) = \mathcal{L}|A\right)=\sum_{t_{i}<t_{l}}\left(1-L(t_{i}-t_{l})\right)q_{n}(t_{i})
\end{align*}
and
\begin{align*} 
    q_{n}(t_{m}) = p_{m}\left(F_{n-1}(t_{m})-F_{n}(t_{m-1})-\Pr\left(Y_{n}(t_{l-1})=\mathcal{L}|A\right)\right).
\end{align*}
Therefore, as before, given $F_{n-1}(t_{m})$ and $q_{n}(t_{i})$
for all $i<m$, we can propagate and calculate $q_{n}(t_{m})$ and
$F_{n}(t_{m})$.

\begin{proof}
We start with an equation equivalent to Eq. \ref{eq:conditional on previous}, 
\begin{align*}
& \Pr\left(Y_{n}(T)=\mathcal{\mathcal{I^{+}}}|A\right)  =&\\
& \Pr\left(Y_{n}(t_{m})=\mathcal{\mathcal{\mathcal{I^{+}}}}|Y_{n-1}(t_{m})=\mathcal{\mathcal{I}},A\right)\Pr\left(Y_{n-1}(t_{m})=\mathcal{\mathcal{I}}|A\right)  =&\\
& \Pr\left(Y_{n}(t_{m})=\mathcal{I^{+}}|Y_{n-1}(t_{m})=\mathcal{\mathcal{I}},A\right)F_{n-1}(t_{m}) &
\end{align*}
where we kept the definition of $F_{j}(t)$. Therefore, almost identically, 
\begin{align*}
    & \Pr\left(Y_{n}(t_{m})=\mathcal{\mathcal{I^{+}}}|Y_{n-1}(t_{m})=\mathcal{\mathcal{I}},A\right)  =& \\
    & \sum_{l=1}^{m}\Pr\left(\text{\ensuremath{y_{n}}'s infection time is \ensuremath{t_{l}}}|Y_{n-1}(t_{m})=\mathcal{\mathcal{I}},A\right)  =&\\
    &\sum_{l=1}^{m}\Pr\left(Y_{n}(t_{l-1})=\mathcal{H},I_{l},Y_{n-1}(t_{l})=\mathcal{\mathcal{I}}|Y_{n-1}(t_{m})=\mathcal{\mathcal{I}},A\right). &
\end{align*}

Eq. \ref{eq:time-shift} follows up to the last line, where:

\begin{align*}
    & \Pr\left(Y_{n}(t_{l-1})=\mathcal{H},I_{l},Y_{n-1}(t_{l})=\mathcal{\mathcal{I}}|Y_{n-1}(t_{m})=\mathcal{\mathcal{I}},A\right)  =&\\
    & p_{l}\Pr\left(Y_{n}(t_{l-1})=\mathcal{H}|Y_{n-1}(t_{l})=\mathcal{I},A\right)\frac{F_{n-1}(t_{l})}{f_{n-1}(t_{m})}  =&\\
    & p_{l}\left(1-\Pr\left(Y_{n}(t_{l-1})=\mathcal{I^{+}}|Y_{n-1}(t_{l})=\mathcal{I},A\right)\right)\frac{F_{n-1}(t_{l})}{F_{n-1}(t_{m})} &
\end{align*}

and,

\begin{align*}
    &\Pr\left(Y_{n}(t_{l-1})=\mathcal{I^{+}}|Y_{n-1}(t_{l})=\mathcal{I},A\right)  =&\\
    & \frac{\Pr\left(Y_{n}(t_{l-1})=\mathcal{I^{+}}|A\right)-\Pr\left(Y_{n}(t_{l-1})=\mathcal{I},Y_{n-1}(t_{l})=\mathcal{H^{+}}|A\right)}{\Pr\left(Y_{n-1}(t_{l})=\mathcal{I}|A\right)}  =&\\
    & \frac{\Pr\left(Y_{n}(t_{l-1})=\mathcal{I}^{+}|A\right)}{\Pr\left(Y_{n-1}(t_{l})=\mathcal{I}|A\right)}  =&\\
    & \frac{\Pr\left(Y_{n}(t_{l-1})=\mathcal{I}^{+}|A\right)}{F_{n-1}(t_{l} )}.
\end{align*}
To summarize, we obtain:
\begin{align*}
    \Pr\left(Y_{n}(T)=\mathcal{\mathcal{I}^{+}}|S_{r}(0)=\mathcal{I}\right) & =\sum_{l=1}^{m}p_{l}\left(1-\frac{\Pr\left(Y_{n}(t_{l-1})=\mathcal{I}^{+}|A\right)}{F_{n-1}(t_{l})}\right)\frac{F_{n-1}(t_{l})}{F_{n-1}(t_{m})}F_{n-1}(t_{m})\\
     & =\sum_{l=1}^{m}p_{l}\left(F_{n-1}(t_{l})-\Pr\left(Y_{n}(t_{l-1})=\mathcal{I}^{+}|A\right)\right)\\
     & =\sum_{l=1}^{m}p_{l}\left(F_{n-1}(t_{l})-F_{n}(t_{l-1})-\Pr\left(Y_{n}(t_{l-1})=\mathcal{L}|A\right)\right)
\end{align*}

Let us denote the probability density function that an infection occurred during interaction $m$ as 
\begin{equation}
q_{n}(t_{m}) = \Pr\left(Y_{n}(t_{m}) = \mathcal{I}^{+}|A\right)-\Pr\left(Y_{n}(t_{m-1}) = \mathcal{I}^{+}|A\right).
\end{equation}
We have, 
\begin{align*}
    q_{n}(t_{m}) =     p_{m}\left(F_{n-1}(t_{m})-F_{n}(t_{m-1})-\Pr\left(Y_{n}(t_{l-1}) = \mathcal{L}|A\right)\right).
\end{align*}
The transition from the latent state to the infected state follows:
\begin{equation}
F_{n}(t_{l}) = \Pr\left(Y_{n}(t_{l}) = \mathcal{I}|A\right) = \sum_{t_{i}<t_{l}}L(t_{i}-t_{l})q_{n}(t_{i})
\end{equation}
while
\begin{equation}
\Pr\left(Y_{n}(t_{l}) = \mathcal{L}|A\right)  =  \sum_{t_{i}<t_{l}}\left(1-L(t_{i}-t_{l})\right)q_{n}(t_{i}).
\end{equation}
Therefore, given $F_{n-1}(t_{m})$ and $q_{n}(t_{i})$ for all $i<m$,
we can propagate and calculate $q_{n}(t_{m})$ and $F_{n}(t_{m})$.
\end{proof}

\end{document}


\title{Supplementary Material\\
    Controlling Graph Dynamics with RL and GNN}
\date{\vspace{-2ex}}
\maketitle
\appendix
\section{Large-scale experiments}
We extend the experiments presented in the paper, conducted on graphs with 200 nodes, to larger graphs. Specifically, here we compare the two best algorithms RLGN (\#8) and SL+GNN (\#4), using graphs with various sizes, from 300 nodes to 1000 nodes.

Table \ref{tab:300-nodes-pa} compares  RLGN with the SL+GNN algorithm on preferential attachment (PA) networks (mean degree $=2.8$). We provide results for various sizes of initial infection  $i_{0}$ and number of available tests $k$ at each step. The experiments show that there is a considerable gap between the performance of the RL and the second-best baseline.
Furthermore, RLGN achieves better performance than the SL+GNN algorithm with 40\%-100\% more tests. Namely, it increases the effective number of tests
by a factor of $1.4X-2X$.

\begin{table}
\centering
    \setlength{\tabcolsep}{3pt}
    \begin{tabular}{|l|c|c|c|c|c|c|}
    \hline 
    \multirow{2}{*}{\textbf{$n=300$}} & \multicolumn{2}{c|}{Init. infection size 5\%} & \multicolumn{2}{c|}{Init. infection size 7.5\%} & \multicolumn{2}{c|}{Init. infection size 10\%}\\
    \cline{2-7} \cline{3-7} \cline{4-7} \cline{5-7} \cline{6-7} \cline{7-7} 
     & \%healthy & \%contained & \%healthy & \%contained & \%healthy & \%contained\\
    \hline 
    SL, $k=1\%$ & $27\pm2$ & $15\pm5$ & $21\pm2$ & $4\pm2$ & $18\pm1$ & $1\pm1$\\
    \hline 
    SL, $k=1.33\%$ & $41\pm3$ & $37\pm6$ & $27\pm2$ & $12\pm4$ & $24\pm2$ & $6\pm3$\\
    \hline 
    SL, $k=2\%$ & ${66\pm4}$ & $76\pm6$ & ${48\pm3}$ & $55\pm7$ & $37\pm2$ & $32\pm6$\\
    \hline 
    RLGN, $k=1\%$ & $50\pm2$ & ${78\pm7}$ & {$43\pm2$} & ${58\pm1}$ & ${40\pm1}$ & ${48\pm6}$\\
    \hline 
    \end{tabular}
    
    \vspace*{0.2 cm} 

    \begin{tabular}{|l|c|c|c|c|c|c|}
    \hline 
    \multirow{2}{*}{\textbf{$n=500$}} & \multicolumn{2}{c|}{Init. infection size 5\%} & \multicolumn{2}{c|}{Init. infection size 7.5\%} & \multicolumn{2}{c|}{Init. infection size 10\%}\\
    \cline{2-7} \cline{3-7} \cline{4-7} \cline{5-7} \cline{6-7} \cline{7-7} 
     & \%healthy & \%contained & \%healthy & \%contained & \%healthy & \%contained\\
    \hline 
    SL, $k=1\%$ & $24\pm2$ & $7\pm4$ & $20\pm1$ & $2\pm1$ & $19\pm1$ & $0\pm1$\\
    \hline 
    SL, $k=1.6\%$ & $48\pm3$ & $54\pm6$ & $35\pm2$ & $27\pm7$ & $29\pm1$ & $11\pm1$\\
    \hline 
    SL, $k=2\%$ & ${67\pm3}$ & $83\pm5$ & ${46\pm2}$ & $53\pm4$ & $38\pm2$ & $37\pm7$\\
    \hline 
    {RLGN, ${k=1\%}$} & $52\pm1$ & ${97\pm2}$ & ${44\pm2}$ & ${75\pm11}$ & ${42\pm1}$ & ${66\pm6}$\\
    \hline 
    \end{tabular}

    \vspace*{0.2 cm} 

    \begin{tabular}{|l|c|c|c|c|c|c|}
    \hline 
    \multirow{2}{*}{\textbf{$n=1000$}} & \multicolumn{2}{c|}{Init. infection size 5\%} & \multicolumn{2}{c|}{Init. Infection size 7.5\%} & \multicolumn{2}{c|}{Init. infection size 10\%}\\
    \cline{2-7} \cline{3-7} \cline{4-7} \cline{5-7} \cline{6-7} \cline{7-7} 
     & \%healthy & \%contained & \%healthy & \%contained & \%healthy & \%contained\\
    \hline 
    SL, $k=1\%$ & $25\pm2$ & $5\pm3$ & $21\pm1$ & $0\pm1$ & $19\pm1$ & $0\pm0$\\
    \hline 
    SL, $k=1.5\%$ & $42\pm2$ & $49\pm6$ & $30\pm1$ & $10\pm3$ & $27\pm1$ & $4\pm2$\\
    \hline 
    SL, $k=2\%$ & ${66\pm1}$ & $84\pm5$ & ${45\pm2}$ & $59\pm5$ & $37\pm1$ & $30\pm1$\\
    \hline 
    {RLGN, ${k=1\%}$} & $52\pm1$ & ${97\pm2}$ & ${44\pm2}$ & ${75\pm11}$ & ${42\pm1}$ & ${66\pm6}$\\
    \hline 
    \end{tabular}
    \caption{A comparison between RLGN and SL+GNN (the best baseline). RLGN performance is highlighted. The number of additional resources needed to surpass the RLGN performance in a given metric is also highlighted.  In many cases, even using SL+GNN with twice as many resources than RLGN performs worse than RLGN. The evaluation was performed on a preferential attachment network with mean degree $2.8$. The number of nodes is indicated at the top of each table.}
    \label{tab:300-nodes-pa}
\end{table}

We further evaluated how models trained on medium-sized graphs generalize well when performing inference on much larger graphs. We trained RLGN and SL+GNN (three model initializations for each) on a preferential attachment network with $1000$ nodes and evaluated its performance of a network with $50,000$ nodes (with the same mean degree $=2.8$). RLGN successfully contained the epidemic in all $150$ evaluation episodes, while the SL+GNN was unable to block the epidemic even once.
The mean percentile of
healthy nodes at the end of the episode was $51\pm1$ for RLGN, while for the SL+GNN it was only $21\pm2$, a difference of more than $15$ STDs.

\section{Ablation Studies}

\subsection{Robustness to size of initial infection} 
We tested the sensitivity of the results to the relative size of the initial infection.
Table \ref{tab:300-nodes-pa} shows results when 4\% of the the network was initially infected, as well as for 7.5\% and 10\%. The results show that RLGN outperforms the baselines in this wide range of infection sizes.

\subsection{Mapping scores to action distribution.} 

Usually, node scores are converted to a distribution over actions using a softmax. This approach is problematic for our case because node probabilities decay exponentially with their scores, leading to two major drawbacks. It discourages exploration of low-score nodes, and limits sensitivity to the top of the distribution, instead of at the top-k selected. Instead, we define the probability to sample an action $a_{i}$ to be
$\Pr(a_{i})=\frac{x_{i}'}{\sum x_{i}'}$, with
\begin{equation}
  x_{i}'=x_{i}-\min_{i}x_{i}+\epsilon \quad , 
\label{eq: score-to-prob}  
\end{equation}
where $\{x_{i}\}$ be the set of scores and $\epsilon$ a constant. 

The fact that we do not use the exponential functions as in softmax, attenuates the probability differences between low-scoring nodes and high-scoring nodes. Furthermore, the parameter $\epsilon$ controls the initial exploration ratio. In standard DNN initialization schemes (like XAVIER), the initial value of $x_{i}$ is expected to be in [-1,1]. If $\epsilon\gg1$ than the dominant term in \eqref{eq: score-to-prob} is $\epsilon$. This promotes exploration initially, as all actions are likely to be sampled in the early training stages.

\begin{figure}
    \centering
    \includegraphics[width=0.6\columnwidth]{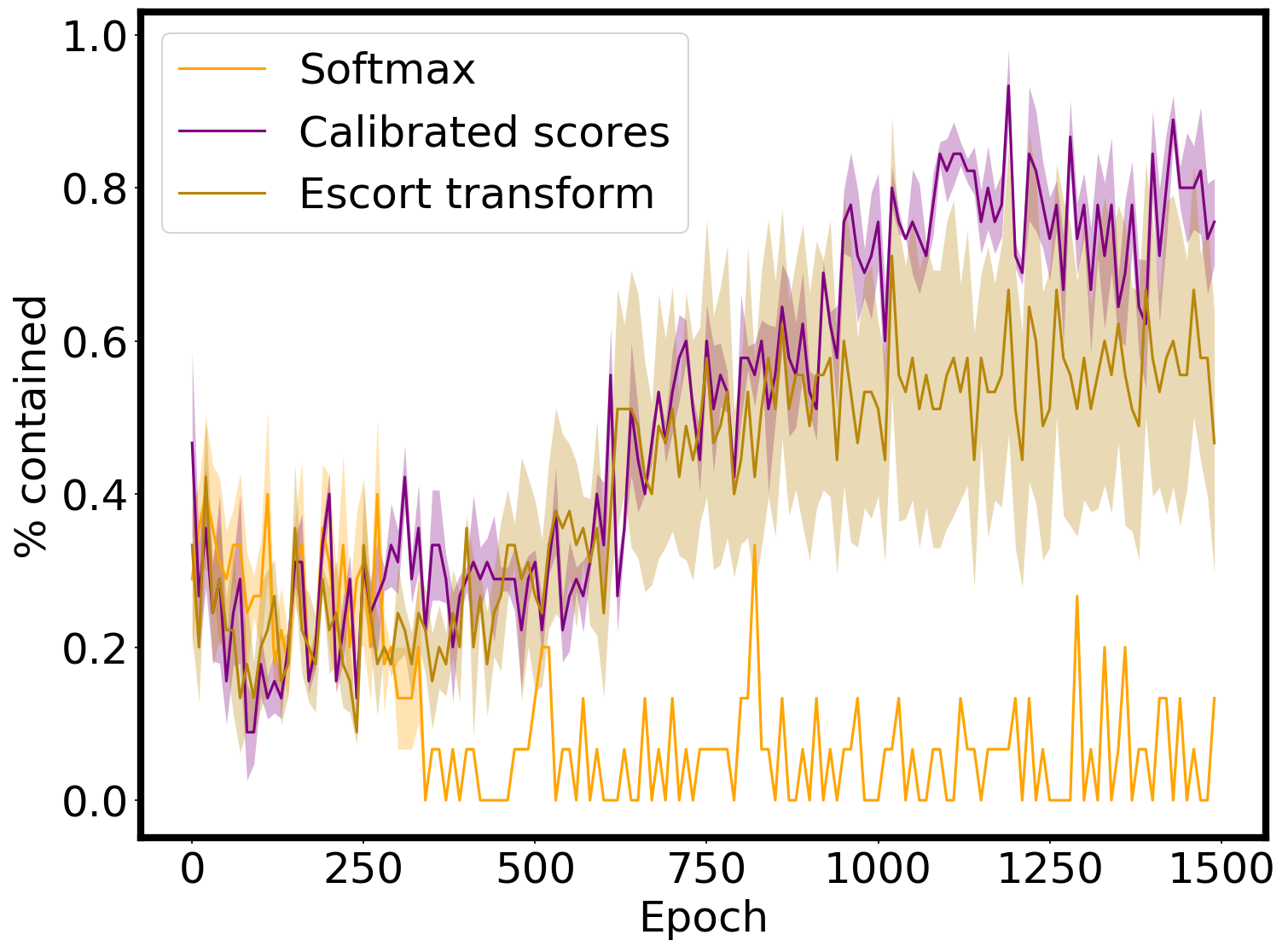}
    \caption{The fraction of contained epidemics during training on a preferential attachment model with $200$ nodes and a mean degree $2.8$. For non-normalized mapping, only one of the three seeds in the softmax distribution simulation completed training due to numerical instability. No stability issues were observed when using  the calibrated scores normalization scheme described by Eq. \eqref{eq: score-to-prob}.
    \label{fig:score-to-prob}}
\end{figure}

\paragraph{Ablation study.} 
We compare the performance of our score-to-probability function (calibrated-scores) to the popular softmax (Boltzmann) distribution. In practice, in most instances, we were unable to train a model using the softmax distribution as the neural network weights diverge. Fig.~\ref{fig:score-to-prob} presents the training curve in one of the few instances that did converge. It is clear that the model was not able to learn a useful policy while using the calibrated-scores probability function resulted in a corresponding value of more than $0.75$. 

\begin{table}
\centering
    \begin{tabular}{|c|c|c|}
        \hline 
         & $\%contained$ & \# training epochs\\
        \hline 
        \hline 
        Sigmoid & $0.84\pm0.05$ & 1210\\
        \hline 
        GRU & $0.91\pm0.03$ & 810\\
        \hline 
        $L_{2}$ norm. & $\mathbf{0.93\pm0.02}$ & \textbf{500}\\
        \hline 
    \end{tabular}
    \caption{\label{tab:norm-table} Training time and fraction of contained epidemic for three normalization schemes. The $L_2$ normalization scheme is fastest and achieves the best performance.}
\end{table}

\subsection{Normalization in scale-free networks.} 

RNN are known to suffer from the problem of exploding or vanishing gradients. This problem is exacerbated in a RNN-GNN framework used for RL algorithms, because they may be applied for arbitrary long episodes, causing internal state to grow unbounded. This problem is particularly severe if the underlying graph contains hubs (highly connected nodes). One approach to alleviate this problem, is by including an RNN like a GRU module, where the hidden state values pass through a sigmoid layer. As the magnitude of the input grows, gradient become smaller and training slows down. We found that the problem can be solved by directly normalize each node hidden state ($L_2$ norm). 

Scale-free network contain high-degree nodes (``hubs") with $O(n)$ neighbors with high probability.  As a simple case, consider a star graph with a large number of nodes. In a GNN framework, it receives updates from a large number of neighbors and its internal state increases in magnitude. At the next time that the GNN module is applied (e.g., next RL step), the growing internal state increases the magnitude of the internal state of its neighbors. This leads to a positive-feedback loop that causes the internal state representation to diverge. Since RL algorithms may be applied for arbitrary long periods, the internal state may grow unbounded unless corrected. 

\paragraph{Ablation experiment.} 
We compared the suggested normalization to a number of other alternative normalization methods. (1) Applying a sigmoid layer after the hidden state update module $G$. (2) Replace the hidden state update module with a GRU layer. (3) Apply $L_2$ normalization to each feature vector $h_v(t)$ (similarly to \cite{Hamilton2017}) (4) Normalize the feature vector matrix by its $L_2$ norm.
These four normalization schemes span three different types of normalization: single-feature normalization (1+2), vector normalization (3), and matrix normalization (4).

Table \ref{tab:norm-table} presents the score after training and the number of training steps required to achieve it (see supplementary information for definitions). Method (4) was unstable and the training did not converge, therefore it was omitted from the table.  The main reason for the training time difference is that without normalization, the DNN weights' magnitude increases. In a GRU module, or with a direct application of a sigmoid layer, the features pass through a sigmoid activation function. When the  magnitude of the input to this layer is large, the gradient is very small due to the sigmoid plateau. This substantially slows down the learning process. 

\subsection{Information processing module. } Our experiments showed the information module has a critical role in improving the performance of the RL-GNN framework. We  performed an ablation study by removing it completely from our DNN module, keeping only the epidemic module. The full DNN module achieved a contained epidemic score of $0.77\pm0.06$, while the ablated DNN module corresponding score was $0.62\pm0.10$, a degradation of more than $20\%$.

\section{Implementation and experiment details}

\subsection{Evaluation metrics}
The end goal of quarantining and epidemiological testing is to minimize the spread of the epidemic. As it is unreasonable to eradicate the epidemic using social distancing alone, the hope is to ``flatten the curve'', namely, to slow down the epidemic progress. To measure the probability of containing the epidemic, we evaluated the percentage of scenarios in which the epidemic spanned less than $\alpha n$ nodes, where $\alpha n$ can represent the public health system capacity. 

In the 2-community setup, where each community has half of the nodes, a natural choice of $\alpha$ is slightly greater than $0.5$, capturing those cases where the algorithm contains the epidemic within the infected community. In all the experiments we set $\alpha=0.6$. The only exception is the three-communities experiments, in which we set the bar slightly higher than $1/3$, and fixed $\alpha=0.4$.

\subsection{Network architecture}
The architecture of the ranking module  is shared by algorithms \#4, \#6 and \#8 with slight variations indicated below.

\paragraph{Input.} We encode the dynamic node features $\zeta_{v}^{d}(t)$
as a one hot vector of dimension $4$. Each of the first three elements
corresponds to one of the three mutually exclusive options, which depends
on the action and node state in the previous step: untested, tested
positive, tested negative. The last entry indicates whether a node
was found positive in the past, namely, if it is quarantined and disconnected
from the graph. The static node features, $\zeta_{v}^{s}(t)$, are
as described in the main paper, topological graph centralities (betweenness,
closeness, eigenvector, and degree centralities) and random node features. The graph centralities were calculated using NetworKit. 
The total number of node features is 9.

\paragraph{Epidemic GNN. }This module $M_{e}$ is composed of a single
graph convolutional layer. The input features are  the last time
step node features. The number of output features is 64. 

\paragraph{Information GNN. }Each message passing module $M^{l}$ contains one hidden layer, where the number of hidden features is 64. After both the hidden and last layer we apply a leaky ReLu layer with leakage constant $0.01$. After aggregating the result using  the addition aggregation function, we apply an additional MLP with one layer (linear+ReLu) on the resulting feature vector. The number of output features is 64. 

We experimented with the numbers of stacked modules $l$ (layers). We found that $l=3$ performed slightly better than $l=2$ but training was considerably slower because the batch size had to be reduced. We therefore used $l=2$ in all experiments reported.

\paragraph{Hidden state update.} The hidden state MLP $G$ is composed of a single linear layer follows by a ReLu activation layer. To keep the resulting hidden feature vector (of dimension 64) norm under check, an additional
normalization scheme is then applied. This module
was replaced by a GRU layer in the ablation studies.

\paragraph{Output layer.} The last module is a single linear layer, with an output dimension as the number of the nodes in the graph.

\paragraph{Learning framework}. We used Pytorch \citep{paszke2017automatic} and Pytorch Geometric  \citep{fey2019fast} to construct the ranking module. We used ADAM with default parameters as our optimizer.

\subsection{Training protocol}
As mentioned in  the main paper, we train the RL and SL by generating random networks and initializing each network by selecting for each instance a random subset of $m_{0}$  infected nodes. We propagate the epidemic until it spans at least $i_0$ infected nodes (for at least $t_0$ steps),
and randomly detect a subset of the infected nodes of size $k_0<i_0$.
At each step, in all algorithms but RL, we pick the top $k$ rated nodes. Each of these nodes is tested, and if detected is positive it is effectively removed from the graph. Otherwise, it is not modified. In RL, we perform the same procedure during the evaluation phase, while during training we sample $k$ nodes using the score-to-probability distribution. 

Each model was training for at most 1500 episodes, but usually, training was completed after 1000 episodes. Each episode contained 1024 steps,
collected by 4 different workers. As our network contains a recurrent module, we propagate each sample in the buffer for three steps, in
a similar fashion to R2D2.

For each setup we described, at least three models were trained using different seeds, and the results are the average over the performance of all models. The errors are the standard deviation of the mean. over at least 100 evaluation episodes for each model.

Each episode lasted for 25 steps, each corresponds conceptually to a day. The transition time from the latent to the infectious state was normally distributed with a mean of two steps and a standard deviation
of 1 step, corresponding to real-world values. The advantage was calculated using the Generalized Advantage framework with parameters $\gamma=0.99,\lambda=0.97.$

Table \ref{table:parameters}  presents the simulation parameters used in the main paper.
We shall make the repository and code available online.

\begin{table}
\centering
    \begin{tabular}{|l|c|}
    \hline 
    minimal \#propagation steps ($t_{0}$) & 4\\
    \hline 
    \multirow{3}{*}{minimal \#infected component size ($i_{0}$)} & communities: 4 (same community)\\
    \cline{2-2} 
     & preferential attachment: 5\%\\
    \cline{2-2} 
     & contact tracing: 7\%\\
    \hline 
    Learning rate & $3\cdot10^{-4}$\\
    \hline 
    $\lambda$ & 0.97\\
    \hline 
    $\gamma$ & 0.99\\
    \hline 
    Entropy loss weight & 0.01\\
    \hline 
    Value loss weight & 0.5\\
    \hline 
    Batch size & 256 (128 if \#nodes>200)\\
    \hline 
    3-communites SBM matrix & $\left(\begin{array}{ccc}
    0.6 & 0.001 & 0\\
    0.001 & 0.6 & 0.001\\
    0 & 0.001 & 0.6
    \end{array}\right)$\\
    \hline 
    2-communites SBM matrix & $\left(\begin{array}{cc}
    0.6 & 0.0022\\
    0.0022 & 0.6
    \end{array}\right)$\\
    \hline 
    \end{tabular}
    \caption{Parameters table}
    \label{table:parameters}
\end{table}

\section{Complex Network datasets}

\subsection{Community-based graphs}
The Stochastic Block Model (SBM) is defined by (1) A partition of nodes to $m$ disjoint communities $C_{i}$, $i=1\dots m$; and (2) a matrix $P$ of size $m\times m$, which represents the edge probabilities between nodes in different communities, namely, the matrix entry $P_{i,j}$ determines the probability of an edge $(v,v')$ between $v\in C_{i}$ and $v'\in C_{j}$. The diagonal elements in $P$ are often much larger than the off-diagonal elements, representing the dense connectivity in a community, compared to the intra-connectivity between communities.

\begin{figure}
    \centering
   \begin{tabular}{cc}
       \includegraphics[width=0.5\columnwidth]{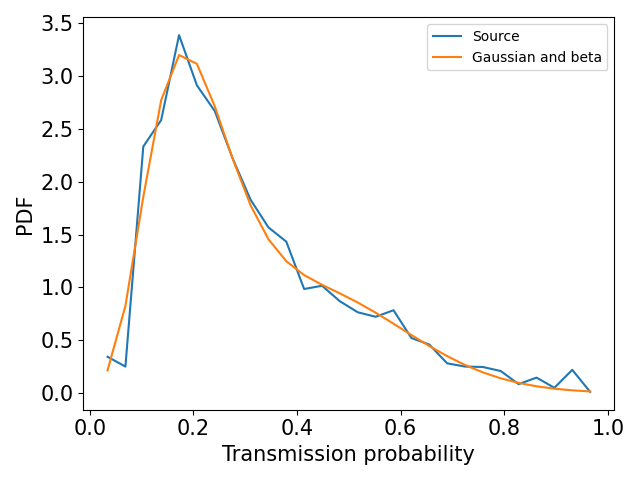}& \includegraphics[width=0.5\columnwidth]{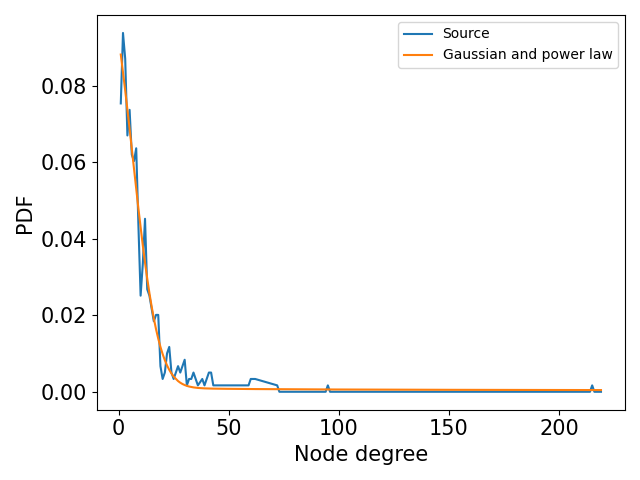}\\
       (a) & (b)
       \end{tabular}
       \caption{Statistics of a real-world contact-tracing graph. (a) The empirical transition probability $P(p_e)$ on a contact tracing network and our suggested curve fit. (b) The degree distribution on the contact tracing network, along with its fit.}
       \label{fig:transmission_prob}
\end{figure}

\subsection{Contact-tracing graphs}
We used anonymized information about a real-world contact tracing effort collected by the health authorities of our country. Fig. \ref{fig:transmission_prob}(a)
presents the degree distribution in this data, and the transmission probability is presented in Fig. \ref{fig:transmission_prob}(b). The latter was derived based on the contact properties, such as the length
and the proximity of the interaction. On average, $1.635\pm0.211$ interactions with a significant transmission probability were recorded per-person per-day. We generated random networks based on these distributions using a configuration model framework \citep{Newman2010a}. The fitted model for the degree distribution is a mixture of a Gaussian and a power-law distribution
\begin{align}
    P(degree=x)=0.47\cdot\mathcal{N}(0.41,0.036)+0.53\cdot Beta(5.05,20.02).    
\end{align}

The fitted model for the transmission probability is a mixture of a Gaussian and a Beta distribution
\begin{align}
    P(p_{e}=x)=2.68\cdot\mathcal{N}(-4.44,11.18)+3.2\cdot10^{-3}\cdot x^{-0.36}.
\end{align}


\section{The rational behind out RL Framework}
Since the action space is combinatorial and huge, action-value methods are prohibitively hard to learn. Instead, we pose the problem as a ranking
problem. Our models process the partial observation $$O(t)=\left\{ P(t=0),\left\{ G(t')|t'<t\right\} ,\left\{ T(t')|t'<t\right\} \right\},$$
and output a score over $k$ nodes. During inference, we select the top-$k$ highest score nodes. During training, we convert the score to probabilities using a score-to-probability mapping, and sample $k$ nodes from the resulting probability distribution.

\section{The tree model}
We describe here our tree model baseline (algorithm \#1). This baseline assumes that the underlying network is a tree. In this case, we can analytically solve for the probability a node is infected, given that the root of the tree was infected at time $t_0$.

Our goal is to calculate the probability that each node $j$ will
be infected
\[
\Pr\left(ST_{j}(T)\in\left\{ \mathcal{I},\mathcal{L}\right\} \right)
\]

We start with a simple case.

\subsection{Simple case: No latent state}

Let us first consider a simple model in which the epidemic spreads
on a tree like structure with a single epidemic source, a.k.a. patient-zero,
as the root. For now, let us assume there is no latent state. For
every node $j$ there is a single path from the node to the root,
which we shall denote as node $r$. Let us assume the path follows
$\{y_{0}=r,y_{1},y_{2},..y_{n-1},y_{n}=j\}$. Assume that a sequence
of interaction between node $y_{n}$ and $y_{n-1}$ happened during
$[0,T]$ at discrete times $(t_{1},t_{2},...t_{m})$, and each interaction
is characterized by an infection probability $(p_{1},p_{2},...p_{m})$.
We shall evaluate the expression:
\[
F_{n}(T)=\Pr\left(ST_{n}(T)=\mathcal{\mathcal{I}}|ST_{r}(0)=\mathcal{I}\right)
\]

by induction. For abbreviation, we shall write $ST_{y_{i}}(t)=Y_{i}(t)$
and denote the event $ST_{r}(0)=\mathcal{I}$ as $A$. This can analytically
calculated using dynamic programming. In short, the state of node
$n$ at the time of interaction $m$ is a function of its state at
penultimate interaction time $F_{n}(t_{m-1})$, the interaction transmission
probability $p_{m}$, and the predecessor node $n-1$ state at time
$m$, $F_{n}(t_{m-1})$.

\begin{align*}
    F_{n}(t_{m}) & =F_{n}(t_{m-1})+p_{m}\left(F_{n-1}(t_{m})-F_{n}(t_{m-1})\right)\\
     & =p_{m}F_{n-1}(t_{m})+F_{n}(t_{m-1})\left(1-p_{m}\right)
\end{align*}

The first term is the probability to get infected at the $m$ interaction,
and the second term is the probability to get infected before hand.
We set $\Pr\left(Y_{n-}(t_{m})=\mathcal{\mathcal{I}}|A\right)$ We can write the conditional probability using the graphical model decomposition and obtain 
\begin{align}
    & \Pr\left(Y_{n}(T)=\mathcal{\mathcal{I}}|A\right) = & \nonumber \\
    & \Pr\left(Y_{n}(t_{m})=\mathcal{\mathcal{I}}|Y_{n-1}(t_{m})=\mathcal{\mathcal{I}},A\right)\Pr\left(Y_{n-1}(t_{m})=\mathcal{\mathcal{I}}|A\right) =& \label{eq:conditional on previous}\\
    & \Pr\left(Y_{n}(t_{m})=\mathcal{\mathcal{I}}|Y_{n-1}(t_{m})=\mathcal{\mathcal{I}},A\right)F_{n-1}(t_{m})\nonumber 
\end{align}
since if the ancestor node is not in an infectious state, the decedent can not be infected. Denote the indicator that interaction $l$ was able to transmit the epidemic as $I_{l}$. We have,
\begin{align*}
    & \Pr\left(Y_{n}(t_{m})=\mathcal{\mathcal{I}}|Y_{n-1}(t_{m})=\mathcal{\mathcal{I}},A\right) & =\\
    & \sum_{l=1}^{m}\Pr\left(\text{\ensuremath{y_{n}}'s infection time is \ensuremath{t_{l}}}|Y_{n-1}(t_{m})=\mathcal{\mathcal{I}},A\right) & =\\
    & \sum_{l=1}^{m}\Pr\left(Y_{n}(t_{l-1})=\mathcal{H},I_{l},Y_{n-1}(t_{l})=\mathcal{\mathcal{I}}|Y_{n-1}(t_{m})=\mathcal{\mathcal{I}},A\right)
\end{align*}

As, for an infection event to take place at it must be that node $y_{n-1}$ was infected at $t_{l}$, node $y_{n}$ was healthy beforehand, and that the interaction resulted in an infection. We can now write this
as 
\begin{align}
    & \Pr\left(Y_{n}(t_{l-1})=\mathcal{H},I_{l},Y_{n-1}(t_{l})=\mathcal{\mathcal{I}}|Y_{n-1}(t_{m})=\mathcal{\mathcal{I}},A\right) & =\nonumber \\
    & p_{l}\Pr\left(Y_{n}(t_{l-1})=\mathcal{H},Y_{n-1}(t_{l})=\mathcal{\mathcal{I}}|Y_{n-1}(t_{m})=\mathcal{\mathcal{I}},A\right) & =\nonumber \\
    & p_{l}\Pr\left(Y_{n}(t_{l-1})=\mathcal{H},Y_{n-1}(t_{m})=\mathcal{\mathcal{I}}|Y_{n-1}(t_{l})=\mathcal{\mathcal{I}},A\right)\frac{\Pr\left(Y_{n-1}(t_{l})=\mathcal{\mathcal{I}}|A\right)}{\Pr\left(Y_{n-1}(t_{m})=\mathcal{\mathcal{I}}|A\right)} & =\label{eq:time-shift}\\
    & p_{l}\Pr\left(Y_{n}(t_{l-1})=\mathcal{H}|Y_{n-1}(t_{l})=\mathcal{I},A\right)\frac{F_{n-1}(t_{l})}{F_{n-1}(t_{m})} & =\nonumber \\
    & p_{l}\left(1-\Pr\left(Y_{n}(t_{l-1})=\mathcal{I}|Y_{n-1}(t_{l})=\mathcal{I},A\right)\right)\frac{F_{n-1}(t_{l})}{F_{n-1}(t_{m})}\nonumber 
\end{align}

The transition from the first line to the second is due to the independence of the interaction infection probability with the history of the participating parties. The third line is simple Bayes' theorem. If a node is infected at time $t_{l}$, it will be infected later on at $t_{m}$, as expressed in line 4. The last line is the complete probability formula. 

We can rewrite $\Pr\left(Y_{n}(t_{l-1})=\mathcal{I}|Y_{n-1}(t_{l})=\mathcal{I},A\right)$ as
\begin{align*}
    & \Pr\left(Y_{n}(t_{l-1})=\mathcal{I}|Y_{n-1}(t_{l})=\mathcal{I},A\right) =& \\
    & \frac{\Pr\left(Y_{n}(t_{l-1})=\mathcal{I}|A\right)-\Pr\left(Y_{n}(t_{l-1})=\mathcal{I},Y_{n-1}(t_{l})=\mathcal{H}|A\right)}{\Pr\left(Y_{n-1}(t_{l})=\mathcal{I}|A\right)} = & \\
    & \frac{\Pr\left(Y_{n}(t_{l-1})=\mathcal{I}|A\right)}{\Pr\left(Y_{n-1}(t_{l})=\mathcal{I}|A\right)} =& \\
    & \frac{F_{n}(t_{l-1})}{F_{n-1}(t_{l})}
\end{align*}
The transition from the first line to the second line is a complete probability transition. The third line is due to the fact that is $y_{n-1}$ was not infected at time $t_{l}$, clearly $y_{n}$ could not be infected before $t_{l}$. We have
\begin{align*}
    F_{n}(t_{m})=\Pr\left(Y_{n-1}(t_{m})=\mathcal{\mathcal{I}}|A\right) & =\sum_{l=1}^{m}p_{l}\left(1-\frac{F_{n}(t_{l-1})}{F_{n-1}(t_{l})}\right)\frac{F_{n-1}(t_{l})}{F_{n-1}(t_{m})}F_{n-1}(t_{m})\\
     & =\sum_{l=1}^{m}p_{l}\left(F_{n-1}(t_{l})-F_{n}(t_{l-1})\right)
\end{align*}

Therefore, given $F_{n-1}(t_{l})$ for all $l\in\{1..n-1\}$ and $F_{n}(t_{l})$
for all $l\in\{1..n\}$, we can directly calculate the infection probabilities, given the initial condition: $F_{i}(0)=\delta_{i,0}$.

We can write the partial density function of $F_{i}(t_{l})$ as $f_{i}(t_{l}) = F_{i}(t_{l})-F_{i}(t_{l-1})$, and obtain: $
f_{n}(t_{m}) = p_{m}\left(F_{n-1}(t_{m})-F_{n}(t_{m-1})\right)$. This allows us to write this with an intuitive formulation
\begin{align*}
    F_{n}(t_{m}) & =F_{n}(t_{m-1})+p_{m}\left(F_{n-1}(t_{m})-F_{n}(t_{m-1})\right)\\
     & =p_{m}F_{n-1}(t_{m})+F_{n}(t_{m-1})\left(1-p_{m}\right)
\end{align*}
The first term is the probability to get infected at the $m$ interaction, and the second term is the probability to get infected before hand.

\subsection{Full analysis with latent states}
The main difference is that the complement of the infectious state
is composed of two states, healthy $\mathcal{H}$, latent $\mathcal{L}$.
We shall designated all the non-infecting states as $\mathcal{H}^{+}=\{\mathcal{\mathcal{H}},\mathcal{L}\}$
and all the infected states as $\mathcal{I}^{+}=\{\mathcal{\mathcal{I}},\mathcal{L}\}$,
and sometime abuse the notation by writing $S_{i}(t)=\mathcal{H}^{+}$.
The transmission from the latent to infectious state follows is a
random variable $L(\tau)$. 

As before, we are interested in the probability that 
\[
\Pr\left(Y_{n}(T)=\mathcal{\mathcal{I}^{+}}|S_{r}(0)=\mathcal{I}\right)
\]
The derivation below shows that, similar to the previous case, we
can solve for this probability using dynamic programming. We have 
\begin{align*}
    \Pr\left(Y_{n}(T)=\mathcal{\mathcal{I}^{+}}|ST_{r}(0)=\mathcal{I}\right) & =\sum_{l=1}^{m}p_{l}\left(F_{n-1}(t_{l})-F_{n}(t_{l-1})-\Pr\left(Y_{n}(t_{l-1})=\mathcal{L}|A\right)\right)
\end{align*}
with $\Pr\left(Y_{n}(t_{l}) = \mathcal{L}|A\right)=\sum_{t_{i}<t_{l}}\left(1-L(t_{i}-t_{l})\right)q_{n}(t_{i})$,  
and $ q_{n}(t_{m}) = p_{m}\left(F_{n-1}(t_{m})-F_{n}(t_{m-1})-\Pr\left(Y_{n}(t_{l-1})=\mathcal{L}|A\right)\right)$. Therefore, as before, given $F_{n-1}(t_{m})$ and $q_{n}(t_{i})$
for all $i<m$, we can propagate and calculate $q_{n}(t_{m})$ and
$F_{n}(t_{m})$.
The modified Eq. \ref{eq:conditional on previous} is 
\begin{align*}
& \Pr\left(Y_{n}(T)=\mathcal{\mathcal{I^{+}}}|A\right)  =&\\
& \Pr\left(Y_{n}(t_{m})=\mathcal{\mathcal{\mathcal{I^{+}}}}|Y_{n-1}(t_{m})=\mathcal{\mathcal{I}},A\right)\Pr\left(Y_{n-1}(t_{m})=\mathcal{\mathcal{I}}|A\right)  =&\\
& \Pr\left(Y_{n}(t_{m})=\mathcal{I^{+}}|Y_{n-1}(t_{m})=\mathcal{\mathcal{I}},A\right)F_{n-1}(t_{m}) &
\end{align*}
where we kept the definition of $F_{j}(t)$. Therefore, almost identically, 
\begin{align*}
    & \Pr\left(Y_{n}(t_{m})=\mathcal{\mathcal{I^{+}}}|Y_{n-1}(t_{m})=\mathcal{\mathcal{I}},A\right)  =& \\
    & \sum_{l=1}^{m}\Pr\left(\text{\ensuremath{y_{n}}'s infection time is \ensuremath{t_{l}}}|Y_{n-1}(t_{m})=\mathcal{\mathcal{I}},A\right)  =&\\
    &\sum_{l=1}^{m}\Pr\left(Y_{n}(t_{l-1})=\mathcal{H},I_{l},Y_{n-1}(t_{l})=\mathcal{\mathcal{I}}|Y_{n-1}(t_{m})=\mathcal{\mathcal{I}},A\right) &
\end{align*}

So Eq. \ref{eq:time-shift} follows up to the last line, where:

\begin{align*}
    & \Pr\left(Y_{n}(t_{l-1})=\mathcal{H},I_{l},Y_{n-1}(t_{l})=\mathcal{\mathcal{I}}|Y_{n-1}(t_{m})=\mathcal{\mathcal{I}},A\right)  =&\\
    & p_{l}\Pr\left(Y_{n}(t_{l-1})=\mathcal{H}|Y_{n-1}(t_{l})=\mathcal{I},A\right)\frac{F_{n-1}(t_{l})}{f_{n-1}(t_{m})}  =&\\
    & p_{l}\left(1-\Pr\left(Y_{n}(t_{l-1})=\mathcal{I^{+}}|Y_{n-1}(t_{l})=\mathcal{I},A\right)\right)\frac{F_{n-1}(t_{l})}{F_{n-1}(t_{m})} &
\end{align*}

and,

\begin{align*}
    &\Pr\left(Y_{n}(t_{l-1})=\mathcal{I^{+}}|Y_{n-1}(t_{l})=\mathcal{I},A\right)  =&\\
    & \frac{\Pr\left(Y_{n}(t_{l-1})=\mathcal{I^{+}}|A\right)-\Pr\left(Y_{n}(t_{l-1})=\mathcal{I},Y_{n-1}(t_{l})=\mathcal{H^{+}}|A\right)}{\Pr\left(Y_{n-1}(t_{l})=\mathcal{I}|A\right)}  =&\\
    & \frac{\Pr\left(Y_{n}(t_{l-1})=\mathcal{I}^{+}|A\right)}{\Pr\left(Y_{n-1}(t_{l})=\mathcal{I}|A\right)}  =&\\
    & \frac{\Pr\left(Y_{n}(t_{l-1})=\mathcal{I}^{+}|A\right)}{F_{n-1}(t_{l} )} 
\end{align*}
To summarize, we obtain:
\begin{align*}
    \Pr\left(Y_{n}(T)=\mathcal{\mathcal{I}^{+}}|S_{r}(0)=\mathcal{I}\right) & =\sum_{l=1}^{m}p_{l}\left(1-\frac{\Pr\left(Y_{n}(t_{l-1})=\mathcal{I}^{+}|A\right)}{F_{n-1}(t_{l})}\right)\frac{F_{n-1}(t_{l})}{F_{n-1}(t_{m})}F_{n-1}(t_{m})\\
     & =\sum_{l=1}^{m}p_{l}\left(F_{n-1}(t_{l})-\Pr\left(Y_{n}(t_{l-1})=\mathcal{I}^{+}|A\right)\right)\\
     & =\sum_{l=1}^{m}p_{l}\left(F_{n-1}(t_{l})-F_{n}(t_{l-1})-\Pr\left(Y_{n}(t_{l-1})=\mathcal{L}|A\right)\right)
\end{align*}

Let us denote the partial density function that an infection occurred during interaction $m$ as $
q_{n}(t_{m}) = \Pr\left(Y_{n}(t_{m}) = \mathcal{I}^{+}|A\right)-\Pr\left(Y_{n}(t_{m-1}) = \mathcal{I}^{+}|A\right)$.
We have, 
\begin{align*}
    q_{n}(t_{m}) =     p_{m}\left(F_{n-1}(t_{m})-F_{n}(t_{m-1})-\Pr\left(Y_{n}(t_{l-1}) = \mathcal{L}|A\right)\right).
\end{align*}
The transition from the latent state to the infected state follows: $
F_{n}(t_{l}) = \Pr\left(Y_{n}(t_{l}) = \mathcal{I}|A\right) = \sum_{t_{i}<t_{l}}L(t_{i}-t_{l})q_{n}(t_{i})$, while $
\Pr\left(Y_{n}(t_{l}) = \mathcal{L}|A\right)  =  \sum_{t_{i}<t_{l}}\left(1-L(t_{i}-t_{l})\right)q_{n}(t_{i})$.
Therefore, given $F_{n-1}(t_{m})$ and $q_{n}(t_{i})$ for all $i<m$,
we can propagate and calculate $q_{n}(t_{m})$ and $F_{n}(t_{m})$.

\bibliographystyle{aaai21}
\bibliography{ref, corona_paper}